%% file: main.tex
\documentclass{article}

\PassOptionsToPackage{x11names}{xcolor}
\usepackage{palatino}

\PassOptionsToPackage{table}{xcolor}
\usepackage[utf8]{inputenc} 
\usepackage[T1]{fontenc}    
\usepackage{geometry}
\usepackage{amsfonts}       
\usepackage{amsmath}
\usepackage{amsthm}
\usepackage{amsbsy}
\usepackage{commath}
\usepackage{siunitx}        
\usepackage{booktabs}       
\usepackage{graphicx}
\usepackage{subfig}
\usepackage{subcaption}
\usepackage{wrapfig}
\usepackage{float}
\usepackage{tabularx}
\usepackage{multirow}
\usepackage{listings}
\usepackage{verbatim} 
\usepackage{enumerate}
\usepackage{bbm}
\usepackage[inline]{enumitem}
\usepackage{mdframed}
\usepackage{titletoc}
\usepackage{xspace}
\usepackage{xcolor}         
\definecolor{papercolor}{HTML}{0668E1}
\definecolor{darkblue}{rgb}{0.0,0.0,0.65}
\definecolor{darkred}{rgb}{0.68,0.05,0.0}
\definecolor{darkgreen}{rgb}{0.0,0.29,0.29}
\definecolor{darkpurple}{rgb}{0.47,0.09,0.29}
\usepackage[USenglish]{babel}
\usepackage{tikz}
\usepackage{microtype}      
\usepackage{hyperref}       
\usepackage{url}            
\usepackage[authoryear]{natbib}
\usepackage[capitalize,noabbrev,nameinlink]{cleveref}
\usepackage[title]{appendix}

\newcommand{\algoname}{\textsc{ContextSSL}\xspace}

\input{math_commands}

\hypersetup{
    colorlinks,
    linkcolor={papercolor!75!black},
    citecolor={papercolor!75!black},
    urlcolor={papercolor!75!black}
}

\lstdefinestyle{mystyle}{
    backgroundcolor=\color{white},
    commentstyle=\color{codegreen},
    keywordstyle=\color{magenta},
    numberstyle=\tiny\color{codegray},
    stringstyle=\color{codepurple},
    basicstyle=\ttfamily\footnotesize,
    breakatwhitespace=false,         
    breaklines=true,                 
    captionpos=b,                    
    keepspaces=true,                 
    numbers=left,                    
    numbersep=5pt,                  
    showspaces=false,                
    showstringspaces=false,
    showtabs=false,                  
    tabsize=2
}
\usepackage{ssArxiv} 

\usepackage{geometry}
\newcommand{\algo}{\textsc{ContextSSL}\xspace}

\newcommand{\sg}[1]{\textcolor{teal}{\textbf{Sharut:}{ #1}}}

\newcommand{\samecolorfootnote}[1]{\textsuperscript{\textcolor{darkred}{#1}}}

\let\oldfootnote\footnote
\renewcommand{\footnote}[1]{\oldfootnote{\textcolor{darkred}{#1}}}

\definecolor{codepurple}{rgb}{0.58,0,0.82}
\colorlet{alternateRowColor}{codepurple!5}

\author{
  Sharut Gupta$^\dagger$\samecolorfootnote{\textsuperscript{*}}, Chenyu Wang$^\dagger$\samecolorfootnote{\textsuperscript{*}},
  Yifei Wang$^\dagger$\samecolorfootnote{\textsuperscript{*}},
  Tommi Jaakkola$^\dagger$ ,
  Stefanie Jegelka$^\ddagger$$^\dagger$  \\
  $^\dagger$MIT CSAIL,  $^\ddagger$TU Munich\\
  \texttt{\{sharut, wangchy, yifei\_w, jaakkola, stefje\}@mit.edu} 
}
\title{In-Context Symmetries: Self-Supervised Learning \\ through Contextual World Models}

\begin{document}
\maketitle

\begin{abstract}
At the core of self-supervised learning for vision is the idea of learning invariant or equivariant representations with respect to a set of data transformations. This approach, however, introduces strong inductive biases, which can render the representations fragile in downstream tasks that do not conform to these symmetries. In this work, drawing insights from world models, we propose to instead learn a general representation that can adapt to be invariant or equivariant to different transformations by paying attention to \emph{context} --- a memory module that tracks task-specific states, actions, and future states. Here, the action is the transformation, while the current and future states respectively represent the input's representation before and after the transformation.  Our proposed algorithm, \textbf{Context}ual \textbf{S}elf-\textbf{S}upervised \textbf{L}earning (\algoname), learns equivariance to all transformations (as opposed to invariance). In this way, the model can learn to encode all relevant features as general representations while having the versatility to tail down to task-wise symmetries when given a few examples as the context. Empirically, we demonstrate significant performance gains over existing methods on equivariance-related tasks, supported by both qualitative and quantitative evaluations.

\end{abstract}

\input{introduction}

\input{method}

\input{results}

\input{related}
\input{conclusion}

\input{acknowledgement}

\clearpage





{
  \bibliographystyle{plainnat}
  \bibliography{references}
}


\appendix
\input{appendix}
\end{document}

%% file: math_commands.tex
\usepackage{amsmath,amsfonts,bm}

\def\eqref#1{equation~\ref{#1}}

\def\1{\bm{1}}

\DeclareMathAlphabet{\mathsfit}{\encodingdefault}{\sfdefault}{m}{sl}
\SetMathAlphabet{\mathsfit}{bold}{\encodingdefault}{\sfdefault}{bx}{n}

\def\gA{{\mathcal{A}}}

\def\gG{{\mathcal{G}}}

\def\sR{{\mathbb{R}}}



%% file: introduction.tex
\section{Introduction}
\footnotetext{Equal contribution.}
Self-supervised learning (SSL) of image representations has made remarkable progress in recent years~\citep{simclr,bardes2021vicreg,ibot,larsson2016learning,gidaris2018unsupervised,bachman2019learning,gidaris2021obow,grill2020bootstrap,shwartz2022pre,misra2020self,chen2020big,he2020momentum,chen2021exploring,zbontar2021barlow,tomasev2022pushing,zhou2022mugs}, achieving competitive performance to its supervised counterparts on various downstream tasks, such as image classification.

\par Most of these works
are based on the joint-embedding architecture (as shown in ~\Cref{fig:fig1}(a)), which encourages the representations of semantically similar (positive) pairs to be close and those of dissimilar (negative) pairs to be more orthogonal. Typically, positive pairs are generated by classic data augmentation techniques that correspond to common pretext tasks, e.g., randomizing color, texture, orientation, and cropping. The alignment of representations for positive pairs can be guided by either invariance~\citep{simclr,bardes2021vicreg,chen2021exploring,he2020momentum,zbontar2021barlow,grill2020bootstrap}, which promotes insensitivity to these augmentations, or equivariance~\citep{gupta2023structuring, devillers2022equimod, dangovski2021equivariant,sie, ijepa, iwm}, which maintains sensitivity to them. However, enforcing invariance or equivariance to a pre-defined set of augmentations introduces strong inductive priors which are far from universal across a range of downstream tasks. For example, invariance to image flipping is useful for image classification but can significantly hurt performance on image segmentation, where retaining sensitivity to flipping is crucial. This often results in brittle representations that necessitate retraining the model with different augmentations tailored to each downstream 
task~\citep{xiao2020should,dangovski2021equivariant}. 

\begin{figure}[!t]
    \centering
    \includegraphics[width=1.\textwidth]{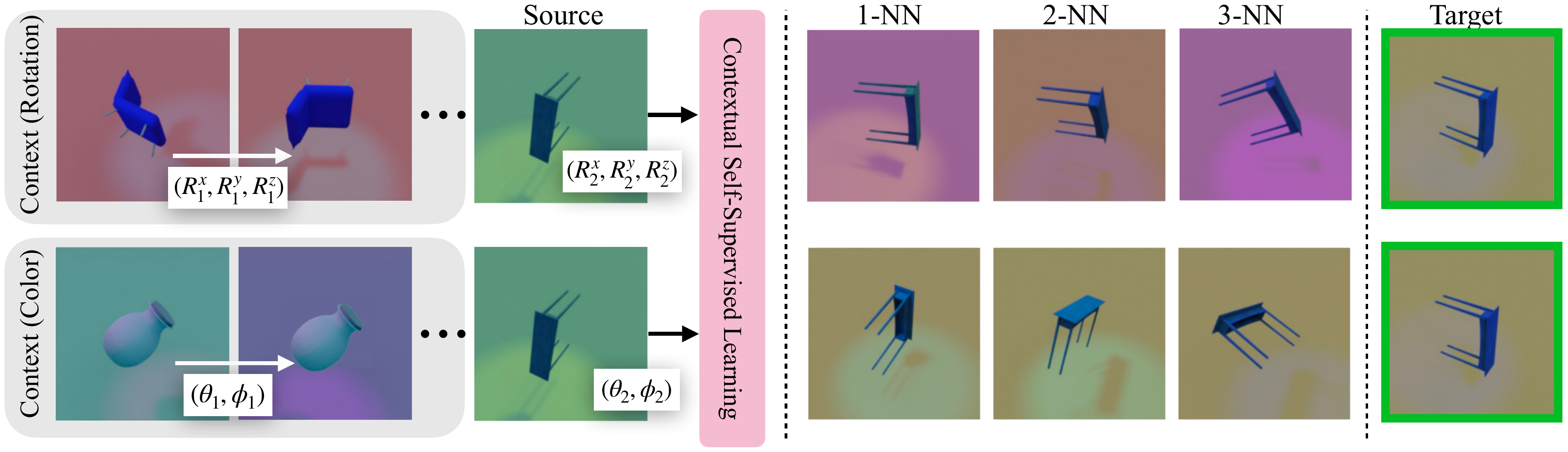}
    \caption{We apply a transformation (rotation or color) on a source image in latent space and retrieve the nearest neighbor (NN) of the predicted representation when the context contains pairs of data transformed by (top row) 3D rotation ($R^x, R^y, R^z$); (bottom row) color transformation $(\theta, \phi)$. In the top row, we see that \algo learns equivariance to rotation and invariance to color as the NN representations match the target's angle but not its color. In the bottom row, it adapts to the color context and enforces the reverse, being equivariant to color and invariant to rotation.} 
    \label{fig:knn_main}
\end{figure}

\par This rigidity of traditional SSL methodologies contrasts sharply with human perceptual abilities, which are highly adaptive, tuning into relevant features based on the \emph{context} of the environment or task at hand. For example, humans focus more on color details when identifying flowers and on spatial orientation, such as rotation angle, when determining the time on analog clocks. It suggests that the required feature invariances or equivariances should also vary across different tasks or contexts, 
which motivates our central question.
\begin{center}
  \emph{Can incorporating context into self-supervised vision algorithms eliminate augmentation-based inductive priors and enable dynamic adaptation to varying task symmetries?}
\end{center}

This work suggests a positive answer to this question by proposing to enhance the current joint 
embedding architecture with a finite context --- an abstract representation of a task, containing a few demonstrations that inform about task-specific symmetries, as shown in~\Cref{fig:fig1}(c). Based on this idea, we propose \textbf{Context}ual \textbf{S}elf-\textbf{S}upervised \textbf{L}earning (\algoname), a contrastive learning framework that uses a transformer module to adapt to selective invariance or equivariance to transformations by paying attention to context representing a task. 
Unlike previous approaches with built-in symmetries, the ability of \algo to adapt to varying data symmetries---all without undergoing any parameter updates---enables it to learn a general representation across tasks, devoid of specific inductive priors.

This unique prospect makes our model a promising approach to building world models~\citep{hafner2019dream, hafner2023mastering,hu2023gaia,sekar2020planning,yang2023learning} for vision. World models are essential for building representations of the world based on past experiences, akin to how humans form their internal world representations. Recently, efforts have been made to adapt world modeling into vision through Image World Models (IWM)~\citep{iwm} (~\Cref{fig:fig1}(b)), that consider transformations as actions and the input and its transformed counterpart as world states at different time steps. However, these approaches also enforce equivariance to a predefined set of actions, such as color jitter. \algoname addresses this challenge by enhancing traditional IWMs with context, a model we refer to as \emph{Contextual World Models}. We demonstrate that in the absence of context, \algoname learns a general representation by encoding all relevant features and data transformations. As the context increases, the model tailors its symmetries to a task, encouraging equivariance to a subset of transformations and invariance to the rest (as shown in~\Cref{fig:knn_main}). This approach promotes learning a general representation that can flexibly adapt to the symmetries relevant to various downstream tasks, eliminating the need to learn separate representations for each task. We empirically validate our approach on the 3D Invariant Equivariant Benchmark (3DIEBench) and CIFAR-10, extending to transformations such as rotations, cropping, and blurring. \\
To summarize, the main contributions of our work are:
\begin{itemize}
  
  \item We propose \algoname, a self-supervised learning algorithm that adapts to task-specific symmetries by paying attention to context. Our method resolves the long-standing challenge of enforcing fixed invariances and equivariances to handcrafted data augmentations, enabling adaptive and task-sensitive representations without parameter updates.

  \item We show that learning with context is prone to identifying shortcuts and subsequently propose two key modules to address it: a context mask and an auxiliary predictor. 
  
  \item We demonstrate the efficacy of our approach on 3D Invariant Equivariant Benchmark (3DIEBench) and CIFAR10, showing its ability to selectively learn invariance or equivariance to transformations such as color and rotation while maintaining similar performance on invariant (classification) benchmarks. We extend \algoname to supervised learning, demonstrating its ability to effectively leverage context to identify features defining a task. 

\end{itemize}

\begin{figure}[!t]
    \centering
    \includegraphics[width=1.\textwidth]{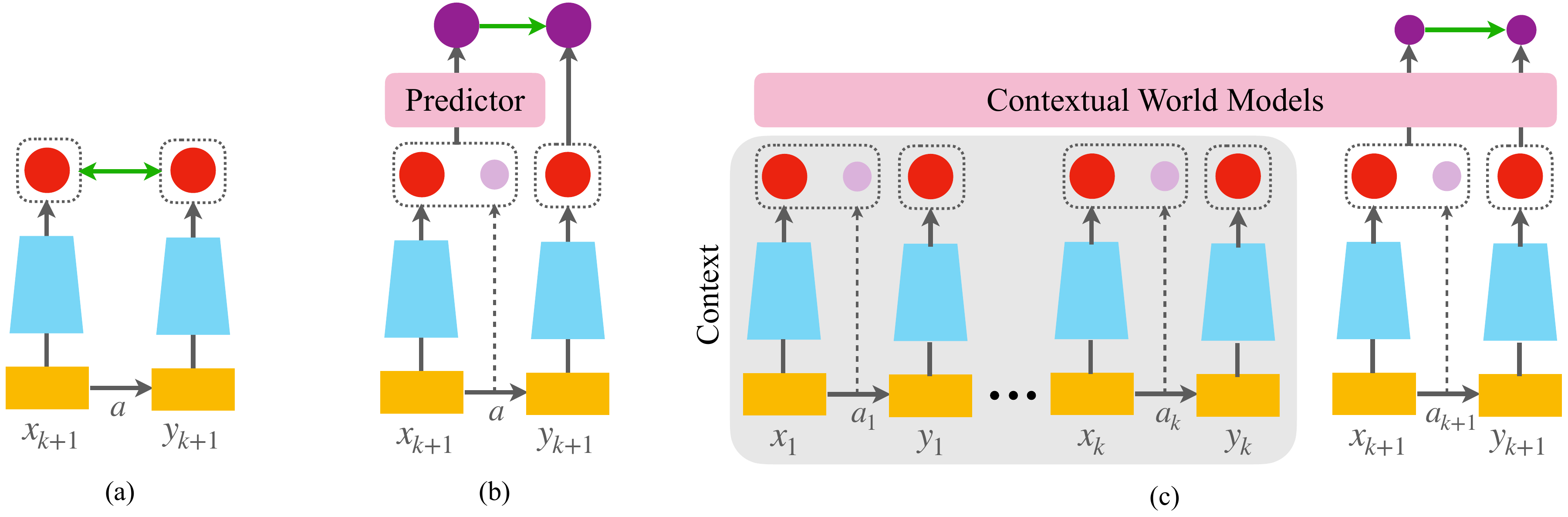}
    \caption{Family of approaches in self-supervised learning (a) \textbf{Joint Embedding} methods~\citep{simclr,bardes2021vicreg,dino} encode invariances to input transformations $a$ by aligning representations across views of the same image; (b) \textbf{Image World Models}~\citep{iwm, ijepa} train a world model in the latent space and encode equivariance to input transformations; (c) \textbf{Contextual World Models} (ours) selectively enforce equivariance or invariance to a subset of input transformations based on context $\{(x_i, a_i, y_i)\}_{i=1}^{k}$}
    \label{fig:fig1}
\end{figure}

%% file: method.tex







\section{Augmentation-based Inductive Bias in Self-Supervised Learning}


The goal of self-supervised learning (SSL) is to derive meaningful data representations without relying on human-labeled data. Given an unlabeled dataset $\mathcal{D}$, SSL methods learn a representation function $f_{\theta}: \mathcal{X} \rightarrow \mathcal{Z}$ that maps input data $x \in \mathcal{X}$ to a latent space $\mathcal{Z}$. 

\subsection{Role of data augmentations in Self-Supervised Learning}
Data augmentations are arguably the most important component in modern SSL methods, where the representation function is learned to map the augmented views of data into latent space. 
The choice of data augmentations plays a crucial role in the quality of the learned representations. Formally, we define an augmentation $A$ as a random variable distributed over a set of $N$ data transformations with domain $\gA=\{a_1,\dots,a_N\}$, where $a_i:\sR^d\to\sR^{d'}$ denotes an input mapping, and $d,d'$ are its input and output dimensions, respectively. 
Among existing SSL methods, there are generally two ways to utilize augmentations, either through invariant learning or equivariant learning. In invariant learning, two random augmentations of the example are drawn, and their representations are pulled together during feature learning to be invariant to the data augmentations as shown in~\Cref{fig:fig1}(a). 
%
%
Instead, in equivariant learning, the features are learned to be sensitive to data
augmentations.\footnotemark\footnotetext{Here, the concept of equivariance is used in a loose sense, meaning that the learned features are sensitive to data augmentations.  Note that since some augmentations are non-invertible (e.g., grayscale), they do not form a group, and exact equivariance is not well-defined.}
%
%
%
Formally, for a representation $Z$, one can use $H(A|Z)$ as a measure of the degree of feature invariance or equivariance: if $H(A|Z)$ is relatively small, the representation $Z$ is nearly equivariant to the augmentation $A$; otherwise, if $H(A|Z)$ is very large (close to $H(A)$), $Z$ is invariant to $A$.
Recent SSL methods~\citep{gupta2023structuring, sie, park2022learning, devillers2022equimod, dangovski2021equivariant} have shown that enforcing equivariance can often lead to better representations compared to enforcing invariance, for two key reasons: 1) Invariance restricts the expressive power of the features learned as it removes information about features or transformations that may be relevant in fine-grained tasks~\citep{lee2021improving,xie2022should}; 2) contrastive learning benefits from partial invariance through implicit equivariance of the projection head~\citep{jing2021understanding}. 

\subsection{Drawbacks of Hardcoding Symmetries in Self-Supervised Pretraining}
As discussed above, a common theme in existing SSL methods is to enforce invariance or equivariance to a specific set of augmentations $A$. For instance, in SimCLR, $A$ is chosen to be a manually selected set of random augmentations such as random cropping, flipping, and color jitter. Therefore, the learned representations, either invariant or equivariant to these augmentations, are tailored to the specific symmetry imposed during pretraining. 
However, in real-world scenarios, no single symmetry is universally applicable across all tasks. For example, object recognition (e.g., a chair) often requires invariance to image color, while certain tasks, e.g., flower recognition, need sensitivity to color information instead. 
Either to include or not to include color information as part of the augmentations can lead to suboptimal performance in certain tasks, causing a fundamental dilemma in existing SSL. 
This leads to brittle representations over a range of downstream tasks, as the model needs to be retrained on different augmentations depending on the downstream tasks, as consistently observed in previous works \cite{xiao2020should,dangovski2021equivariant}.


\section{Beyond Built-in Symmetry: Contextual Self-Supervised Learning}
Recognizing the limitations of existing augmentation-specific SSL methods, we propose a new paradigm: \textbf{Context}ual \textbf{S}elf-\textbf{S}upervised \textbf{L}earning (\algo). Unlike traditional methods, this approach learns a single model that adapts to be either invariant or equivariant based on context-specific augmentations tailored to the needs of the task or data at hand. Instead of enforcing a fixed set of symmetries, \algo learns these symmetries from contextual cues, thus capturing the unique set of features of downstream tasks. This adaptability allows it to serve as a general-purpose SSL framework, capable of learning from a diverse array of pretraining tasks with varying symmetry priors and seamlessly adapting to different downstream tasks. 

To design \algo, we draw inspiration from world modeling~\citep{hafner2019dream, hafner2023mastering, sekar2020planning, yang2023learning}, a widely used framework in reinforcement learning (RL). World modeling aims to build representations of the world from past experience by predicting the next state $x_{t+1}$ from the current state $x_t$ and action $a_t$. This next state prediction task captures the inherent mechanisms of the system and facilitates decision-making. Traditionally applied in RL, the benefits of world modeling in vision have been largely unexplored. Recently, Image World Models (IWM)~\citep{garrido2023self} established a parallel between world models and the image-based SSL by considering data transformations as actions, the representation of input data as world state at time $t$ and that of the transformed input as next world state. 
However, IWMs have two key drawbacks: 1) similar to previous SSL approaches, they rely on a predefined set of data augmentations, such as color, which are not tailored to specific downstream tasks and influence the learned features; 2) they lack the memory module of world models that tracks previous experience in terms of past states, actions, and corresponding next states and provides context to define the current state fully.
\par In light of these ideas and challenges, we model \algo in vision self-supervised learning as \textit{Contextual World Models}. In this way, \algo addresses the key drawbacks of IWMs by 1) encouraging the model to preserve all meaningful features to be able to adapt to symmetry from context and 2) incorporating context to adapt to different task-specific symmetries, removing the need to re-train separate representations for each downstream task. This general ability is akin to human perception, which captures versatile aspects of the input while focusing on specific details depending on the context at hand. 
For instance, humans focus more on color details when identifying flowers and on spatial orientation, such as rotation angle, when determining the time on analog clocks. 

\subsection{Contextual World Models}
Drawing inspiration from the in-context learning~\citep{brown2020language} of foundation models in natural language processing, a natural way to incorporate the memory capabilities of world models is by encoding these abilities as contextual information. In this work, we propose an expressive and efficient implementation of \algo through \textit{Contextual World Models}, where we design a transformer-based module to encode the context and extract contextually equivariant or invariant representations. We begin by baking symmetries in the context --- $(x, a, y)$ using positive pairs $x$ and $y$ transformed by a series of different augmentations.  The key intuition behind our approach is the selective inclusion of augmentation parameters for specific transformation groups: excluding parameters enforces invariance while including them enforces equivariance. This is because providing augmentation parameters allows the model to learn the impact of transformations (equivariance), whereas excluding them during alignment enforces invariance, akin to invariant versus equivariant learning in SSL. We elaborate on these ideas below.



\textbf{Symmetries as Context}. Given a set of groups of input transformations  $\{\gG_1,\dots,\gG_M\}$, the goal of \algo is to build a general representation adaptive to a set of multiple symmetries corresponding to these different groups. For example, each data augmentation, e.g., rotation, translation, as well as their compositions, can serve as different transformation groups. Each group $\gG_c$ can be represented through the joint distribution $P(x, a, y | \gG_c)$, where $x$ is the input sample (sampled from an unlabeled dataset), $a$ represents the parameters of the transformation drawn from $\gG_c$ and applied to $x$, and $y$ is the transformed input. In principle, $x$ can be transformed by a composition of augmentations drawn from multiple transformation groups. For instance, in self-supervised learning, it is common to enrich the learning process by transforming an input image through rotations, crops, and blurring.
In such a case, $a$ represents a subset of the transformation parameters belonging to the group $\gG_c$, applied to $x$ to produce $y$.  We approximate this probability distribution by drawing $K$ samples from the joint distribution and form a context $C(\gG_c)=[(x_1, a_1, y_1), \ldots, (x_K, a_K, y_K)]$, where $x_i, a_i, y_i\sim P(x, a, y | \gG_c),i\in[K]$. 
Therefore, the goal of ContextSSL is to learn data representations $z=f(x,a|C)$ and $z=f(x|C)$ that are adaptive to the data symmetries informed by the context $C$. 
Specifically, our goal is to train representations that become more equivariant to the underlying transformation group $\gG_c$ with increasing context. Further, if $x$ and $y$ are transformed by augmentations from groups apart from $\gG_c$, we aim to learn more invariance to these groups with increase in context $C(\gG_c)$. The degree of equivariance of a representation can be quantified by the error in maintaining consistent transformations. Based on this, a representation $Z$ is considered "more equivariant (invariant)" if it has a lower (higher) error in predicting the transformation parameters i.e. $H(A|Z)$.


\textbf{Contextual World Models.} To implement this broad goal, we propose to adaptively learn the symmetries represented by $\gG_c$ by training the model:
\begin{equation}
    y_{i} \approx h((x_{i}, a_{i}); (x_1, a_1, y_1), \ldots, (x_{i-1}, a_{i-1}, y_{i-1})).
\end{equation} 
While the requested prediction $y_{i}$ concerns only the inputs $x_{i}$ and $a_{i}$, the model can now pay attention to the experience so far, enforcing relevant symmetries for the augmentation group $\gG_c$. The predictor $h$ is updated by minimizing the loss at each context length $\sum_{i=1}^K\ell(h((x_{}, a_{i}); C_{i-1}), y_i)$ where $C_i = \{(x_1, a_1, y_1), \ldots, (x_{i-1}, a_{i-1}, y_{i-1})\}$ represents the context before index $i$.

\par A natural way to facilitate such context-based training is through attention mechanisms in transformer-based autoregressive models. Large language models exhibit a remarkable capability of in-context learning --- the ability to generalize to unseen tasks on the fly merely by paying attention to a few demonstrative examples of the task. \citet{gupta2023context} among others, have leveraged this capability to generalize to different distributions merely by paying attention to unlabeled examples from a domain. Inspired by this, we train a decoder-only transformer model in-context by conditioning on the relevant context $C(\gG_c)$ representing the transformation group $\gG_c$.
%


\subsection{Contextual Self-Supervised Learning (\algo)}
Motivated by the above ideas, we begin by constructing pairs of points $\{(x_i, y_i)_{i=1}^K\}$ by either 1) sampling a transformation group $\gG$ and transforming $x_i$ by augmentation from $\gG$ to $y_i$; or 2) if available, sampling a meta-latent and its transformation parameters as the difference between their individual latent parameters. We use the former construction in datasets such as CIFAR10 but use meta-latents such as 3D pose, lighting etc. for datasets such as 3DIEBench~\citep{sie}. Note that pairs of data can also be transformed by a series of augmentations sampled from other transformation groups. However, as previously discussed, the transformation parameters used in the context $C(\gG)$ of group $\gG$ are solely those of the augmentations belonging to the group. 

Following this, as illustrated in~\Cref{fig:fig1}, each input sample $\{(x_i, y_i)\}_{i=1}^K$ from the context is independently transformed by the encoder into its corresponding latent representation. Next, representations of the input samples $x_i$ are concatenated with their corresponding transformation action $a_i$. This concatenated vector $(x_i, a_i)$  and the representation of the corresponding transformed input $y_i$ collectively form the context corresponding to the symmetry $\gG$. The corresponding output embeddings are then aligned using the InfoNCE loss, which is minimized at each context length. If $a_i$ is set to zero for all tokens in a sequence, \algo enforces invariance to $\gG$, since it aligns $x_i$ and $y_i$ without conditioning on the transformation parameters. Overall, we optimize the following loss:

\begin{equation}\label{eqn: InfoNCE loss}
\mathcal L_{\text{contrastive}}(h) = \mathbb{E}_{\gG \sim \{\gG_1, ...\gG_M\}} \mathbb{E}_{C(\gG)}\sum_{i=1}^K \bigg [ - \log \frac{\text{exp } {h((x_i, a_i) | C_i(\gG))^\top h(y_i| C_i(\gG)) /\tau}}{ \sum_{j=1}^K \text{exp } {h((x_i, a_i)| C_i(\gG))^\top h(y_j| C_j(\gG)) / \tau}}\bigg ]
\end{equation} 
where transformed data tokens $y_j$ ($j\neq i$) form the negatives. We use a similar symmetric loss term using $y_i$ as the anchor, $(x_i, a_i)$ and $(x_j, a_j)$ $(j\neq i)$ as the positive and negatives respectively. 

At inference, we tailor the extraction of representations to match the specific requirements of the downstream task, whether it benefits from equivariance or invariance to a transformation group $\gG$. In particular, if the task benefits from equivariance, we extract the representations of the test data at the maximum context length used during training $K$, by constructing $\{(x_i, a_i, y_i)\}_{i=1}^K$ as its preceding context. Here $a_i$ belongs to the group $\gG$ and is used to transform other unlabelled data from the test set $x_i$ into $y_i$. On the contrary, if the downstream task benefits from invariance to the group, we use $\{(x_i, 0, y_i)\}_{i=1}^L$ as the preceding context. This notion can be generalized to enforce equivariance to a subset of groups and invariance to another.
Specifically, including the augmentation parameters for transformations in a group 
$\gG$ in the context enforces equivariance, while excluding them enforces invariance. In both cases, the data are still transformed using augmentations, regardless of the type of symmetry desired. This flexibility of context creation in \algo allows us to tailor the representations to different symmetries and optimize for the model's performance across various tasks. However, this implementation bears two key challenges, as detailed below. 

\begin{figure}[!htb]
    \centering
    \includegraphics[width=\textwidth]{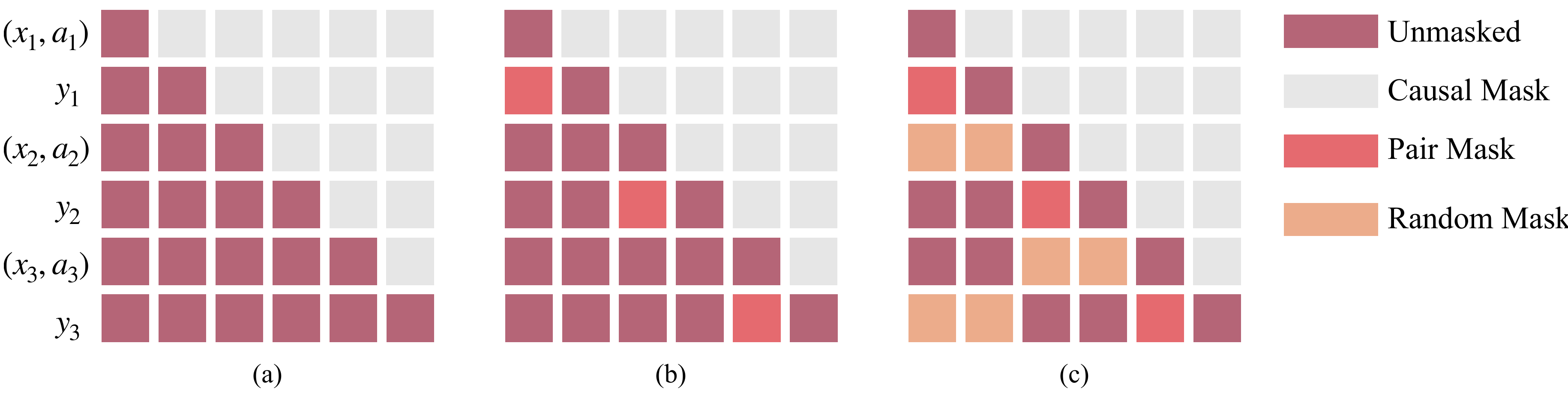}
    \caption{(a) Traditional causal attention mask; (b) corresponding pair masking on top of causal attention to prevent shortcuts when aligning positives; (c) Context Mask used in \algo to prevent shortcuts while distinguishing positives from negatives.
}\label{fig:masking_illustration}
\end{figure}

\noindent \textbf{Context Masking.} Given that $(x_i, a_i)$ precedes $y_i$ in the context sequence, a trivial solution to minimizing the alignment loss arises where the model treats the embeddings of $(x_i, a_i)$ identical to $y_i$ due to its access to $x_i$. This phenomenon, often referred to as shortcut learning, poses a significant challenge as it leads the model to collapse to constant representations for each pair $(x_i, y_i)$, all while perfectly minimizing the loss. We address this challenge by masking out the input token $(x_i, a_i)$ for each token $y_i$ in the context. As a consequence, when encoding the token $y_i$, the transformer only has access to past context $C_i =\{(x_1, a_1, y_1), \ldots, (x_{i-1}, a_{i-1}, y_{i-1})\}$, excluding its corresponding positive sample $(x_i, a_i)$. 
\par This masking approach ensures that both the anchor and its corresponding positive share the same context, thus promoting the alignment of positive samples based on semantic relationships rather than mere replication. However, as shown in~\Cref{fig:fig_ablations_mask} for $p=0$, a residual challenge of shortcut learning persists when distinguishing the positives from the negatives. Since the context corresponding to each negative differs from that of the anchor and the positive, the model could employ trivial solutions, such as using the mean of the context vector to differentiate between positives and negatives.
%
\par To mitigate this issue, we introduce an additional layer of randomness to our masking strategy. Specifically, for each token in the context vector, we implement random masking of the corresponding pairs with a probability $p$ for tokens preceding it. Specifically, if $(x_i,a_i)$ is chosen to be masked out, so is $y_i$ and vice versa. This ensures that for a given anchor token, both the positive and the negatives have different contexts from the anchor, thereby necessitating a deeper, semantic understanding to distinguish the positives from the negatives effectively. 

\noindent \textbf{Avoiding collapse to Invariance.}
A trivial but undesirable solution that minimizes our optimization objective is invariance to the input transformations i.e. the trained model can ignore the transformation parameters and collapse back to behaviors associated with invariance-based methods. 
As illustrated in~\Cref{fig:ablation_aug_pred}, naively training \algoname leads to poor equivariance with respect to the transformations. Previous works~\citep{sie} have also identified this concern and proposed specialized architectures that incorporate transformation parameters directly into the model, thereby outputting the predictor's weights and ensuring effective utilization of these parameters.  For our setting, we introduce a rather simple approach that involves jointly training an auxiliary predictor. This predictor is designed to predict the latent transformations of the target sample $y_i$ from the concatenated input vector $(x_i, a_i)$. 
\par In practice, $x_i$ and $y_i$ are typically created as augmentations $t^x_i$ and $t^y_i$ respectively of a raw image, with $a_i$ denotes the relative transformation between the two. Our predictor predicts this latent transformation of the target sample $t^y_i$ from the input vector $(x_i, a_i)$. We predict $t^y_i$ instead of $a_i$, as $a_i$ is already included in the input token $(x_i, a_i)$. As a consequence, the predictor $g$ optimizes the following mean-squared loss

\begin{equation}
    \mathcal{L}_{\text{predictor}}(g) = \mathbb{E}_{\gG \sim \{\gG_1, ...\gG_M\}} \mathbb{E}_{C(\gG)}\sum_{i=1}^K (t^y_i - g(x_i,a_i))^2
\end{equation}
 
Finally, we propose the following objective that combines our transformation prediction loss with $\mathcal L_{\text{\algo}}$ loss:
\begin{equation}\label{eq: final_loss}
    \mathcal L_{\text{\algo}} = \mathcal L_{\text{contrastive}} + \lambda \mathcal{L}_{\text{predictor}}
\end{equation}
where $\lambda$ weights the transformation prediction loss. Setting $\lambda$ to zero reverts back to invariance to all input transformations, rendering context inconsequential. We use a symmetric loss, alternating between $(x_i,a_i)$ and $y_i$ as the anchor. When $y_i$ is the anchor, the predictor loss is trained to predict $t^y_i$ from $y_i$ and minimizes the loss $\mathbb{E}_{\gG \sim \{\gG_1, ...\gG_M\}} \mathbb{E}_{C(\gG)}\sum_{i=1}^K (t^y_i - g(y_i,0))^2$. In practice, we note that many variations of this approach are possible. For instance, our contrastive loss can also be replaced by other Siamese self-supervised losses such as SimSiam~\citep{chen2021exploring}. Additionally, alternative methods can be used to prevent collapse to invariance. We leave further exploration of these possibilities to future work.








\par

%% file: results.tex
\section{Experimental Results}
To evaluate the efficacy of our proposed algorithm \algoname, our experiments are designed to address the following questions:
\begin{enumerate}
    \item How does \algoname fare against competitive invariant and equivariant self-supervised learning approaches in terms of performance across varying context sizes and different sets of data transformations?
    \item How effectively can \algoname identify task-specific symmetries, both within the scope of self-supervised learning and beyond?
    \item What roles do specific components such as selective masking and the auxiliary latent transformation predictor play in facilitating the learning of general and context-adaptable representations?
\end{enumerate}

\subsection{Quantitative Assessment of Adaptation to Task-Specific Symmetries}
We use the 3D Invariant Equivariant Benchmark (3DIEBench)~\citep{sie} and CIFAR10 to test equivariance and invariance to multiple data transformations. We compare \algo with 
1) VICReg~\citep{bardes2021vicreg} and SimCLR~\citep{simclr} among the invariant self-supervised approaches; 2) EquiMOD~\citep{devillers2022equimod}, SEN~\citep{park2022learning} and SIE~\citep{sie} amongst the equivariant baselines. To discard the performance gains potentially arising from \algoname's transformer architecture, for each approach $\mathcal{N}$, we replace the original projection head or predictor with our transformer model, denoted as $\mathcal{N}^+$.  We then train two variants of $\mathcal{N}^+$: one without any context (denoted as $\mathcal{N}^+$(c=0)) and another with the full context length.
We further test this for all our equivariant baselines on 3DIEBench, we train equivariant approaches to be equivariant to either only 3D rotation, color transformations, or both. We report the test performance on context lengths 0, 2, 14, 30, and 126. We employ linear classification over frozen features to assess the quality of the invariant representations. For the equivariant counterpart, we report $R^2$ on the task of predicting the corresponding transformation. We report $R^2$ performance on the representation after the encoder for baselines. Detailed results for post-predictor embedding performance are presented in~\Cref{table:repr_emb_compare} in the appendix. Additionally, we use Mean Reciprocal Rank (MRR) and Hit Rate at $k$ (H@k) to evaluate the performance of our context predictor.  More details about pretraining algorithms and training setup are provided in~\Cref{sec:exp_setup}. 


\begin{table*}[!htb]
    \centering
    \caption{Quantitative evaluation of learned representations on invariant (classification) and equivariant (rotation prediction, color prediction) tasks. Additional metrics are reported in~\Cref{sec: predictor_vs_rep_supplement} }
    \resizebox{1.\textwidth}{!}{%
    \begin{tabular}{llcccccccccccc}
        \toprule
        $\gG$ & Method & \multicolumn{5}{c}{Rotation prediction ($R^2$)} & &\multicolumn{5}{c}{Color prediction ($R^2$)} & \multicolumn{1}{c}{Classification (top-1)}  \\
        \cmidrule(lr){3-7} \cmidrule(lr){9-13}\cmidrule(lr){14-14}
        & & 0 & 2 & 14 & 30 & 126 & & 0 & 2 & 14 & 30 & 126 & Representation \\
        \midrule
        & \multicolumn{1}{l}{\textit{\textcolor{gray}{Invariant }}}\\
        & SimCLR & \multicolumn{5}{c}{0.506} & & \multicolumn{5}{c}{0.148} & \textbf{85.3} \\
        & SimCLR$^+$(c=0) & \multicolumn{5}{c}{0.478} & &\multicolumn{5}{c}{0.070} & 83.4\\
        & SimCLR$^+$ & \multicolumn{5}{c}{0.489} & &\multicolumn{5}{c}{0.130} & 81.0 \\
        & VICReg & \multicolumn{5}{c}{0.371} & & \multicolumn{5}{c}{0.023} & 76.3 \\
        & VICReg$^+$(c=0) & \multicolumn{5}{c}{0.356} & & \multicolumn{5}{c}{0.062} & 73.3  \\
        \midrule
        & \multicolumn{1}{l}{\textit{\textcolor{gray}{Equivariant }}}\\
        \multirow{3}{*}{\rotatebox[origin=c]{90}{\parbox{0.8cm}{Rotation \\ + Color}}} & EquiMOD & \multicolumn{5}{c}{0.512} && \multicolumn{5}{c}{0.097} & \textbf{82.4}\\
        & SIE & \multicolumn{5}{c}{\textbf{0.629}} && \multicolumn{5}{c}{\textbf{0.973}} & 71.0\\
        & SEN & \multicolumn{5}{c}{0.585} && \multicolumn{5}{c}{0.932} & 80.7\\
        \midrule
        \multirow{4}{*}{\rotatebox[origin=c]{90}{Rotation}} & EquiMOD & \multicolumn{5}{c}{0.512} && \multicolumn{5}{c}{0.097} & \textbf{82.4}\\
        & SIE & \multicolumn{5}{c}{0.671} && \multicolumn{5}{c}{\textbf{0.011}} & 77.3\\
        & SEN & \multicolumn{5}{c}{0.633} && \multicolumn{5}{c}{0.055} & 81.5\\
        & \algo\footnotemark & 0.734 & 0.740 & 0.743 & 0.743 & \textbf{0.744} & & 0.908 & 0.664 & 0.037 & 0.023 & 0.046 & 80.4 \\
        \midrule
        \multirow{4}{*}{\rotatebox[origin=c]{90}{Color}} & EquiMOD & \multicolumn{5}{c}{0.429} && \multicolumn{5}{c}{0.859} & \textbf{82.1}\\
        & SIE & \multicolumn{5}{c}{0.304} && \multicolumn{5}{c}{0.975} & 70.3\\
        & SEN & \multicolumn{5}{c}{0.386} && \multicolumn{5}{c}{0.949} & 77.6\\
        & \algo\footnotemark & 0.735 & 0.614 & 0.389 & 0.345 & \textbf{0.344} && 0.908 & 0.981 & 0.985 & 0.986 & \textbf{0.986} & 80.4 \\        
        \bottomrule
    \end{tabular}
    }
       \label{table:main_table}
\end{table*}
\footnotetext{In Table 1, both the \algo models are the same and the performance is reported depending on whether the context corresponds to rotation or color augmentation group.}

\textbf{Invariant Classification and Equivariant transformation prediction task.}
As shown in~\Cref{table:main_table}, invariant self-supervised learning methods such as SimCLR and VICReg achieve high downstream classification accuracies but underperform in equivariant augmentation prediction tasks. 
EquiMOD persistently maintains its downstream classification accuracy among the equivariant baselines but exhibits improvements in augmentation prediction tasks only when trained to be equivariant to color. 
When trained to be equivariant to only rotation or both rotation and color jointly, it offers no improvement in performance for augmentation prediction compared to SimCLR, which serves as their base. 
SIE and SEN exhibit sensitivity to the trained transformations and remain less sensitive to the others. However, their degree of invariance or equivariance is much worse compared to \algo. Besides, aligning them with different targeted symmetry groups requires retraining the entire model. In contrast, \algo exhibits equivariance to both rotation and color in the absence of context.
SIE and SEN learn to be equivariant to the transformation they are trained to be equivarient to, and invariant to the other transformation. However, they require retraining of the whole model to align with different targeted symmetry groups. 
Unlike the baseline approaches, which train a separate model for each equivariance, \algo trains a single model that learns equivariance to rotation and invariance to color (or vice versa) depending on the type of context.
As seen from the two rows corresponding to \algo in~\Cref{table:main_table}, when the context corresponds to pairs of data with transformations sampled from the rotation (color) group, the model adaptively learns to be invariant to color (rotation) while improving equivariance to rotation (color). ~\Cref{sec:single env} shows that \algo 
learns equivariance or invariance to the same transformation based on the context.

\begin{table*}[!htb]
    \centering
    \caption{Quantitative evaluation of learned predictors equivariant to only rotation based on Mean Reciprocal Rank (MRR) and Hit Rate H@k on the validation dataset. \algo learns to be more equivariant to rotation with context.}
    \resizebox{1.\textwidth}{!}{%
    \begin{tabular}{lccccccccccccccccc}
        \toprule
        Method & \multicolumn{5}{c}{MRR ($\uparrow$)} && \multicolumn{5}{c}{H@1 ($\uparrow$)} && \multicolumn{5}{c}{H@5 ($\uparrow$)} \\
        \cmidrule(lr){2-6} \cmidrule(lr){8-12}\cmidrule(lr){14-18}
        & 0 & 2 & 14 & 30 & 126 && 0 & 2 & 14 & 30 & 126 && 0 & 2 & 14 & 30 & 126 \\ 
        \midrule
        EquiMOD & \multicolumn{5}{c}{0.16} && \multicolumn{5}{c}{0.05} && \multicolumn{5}{c}{0.22} \\
        SEN & \multicolumn{5}{c}{0.17} && \multicolumn{5}{c}{0.05} && \multicolumn{5}{c}{0.22} \\
        \algo & 0.240 & 0.270 & 0.373 & 0.396 & \textbf{0.402} && 0.108 & 0.129 & 0.223 & 0.245 & \textbf{0.292} && 0.366 & 0.412 & 0.541 & 0.561 & \textbf{0.568}\\
        \bottomrule
    \end{tabular}
    }
\label{table:mrr}
\end{table*}

\textbf{Equivariant Measures Based on Nearest Neighbours Retrieval.} \Cref{table:mrr} illustrates the performance of \algo on MRR and H@k compared to baseline methods with trained equivariance to rotation. \algo outperforms the baseline models, and its performance on all the metrics consistently improves with increasing context length, showing adaptation to rotation-specific features. To put these numbers into perspective, a H@1 score of 0.29 for \algo signifies that the first nearest neighbor is the target embedding 29\% of the time. In contrast, this occurs only 5\% of the time for EquiMod and SEN, which is marginally better than the 2\% expected by random chance. Notably, \algo surpasses the baseline performances even with zero context, demonstrating its ability to learn equivariance without any contextual information.

\subsection{Role of Context Mask and Auxiliary Predictor}

\begin{figure}[!htb]
    \centering
    \includegraphics[width=\textwidth]{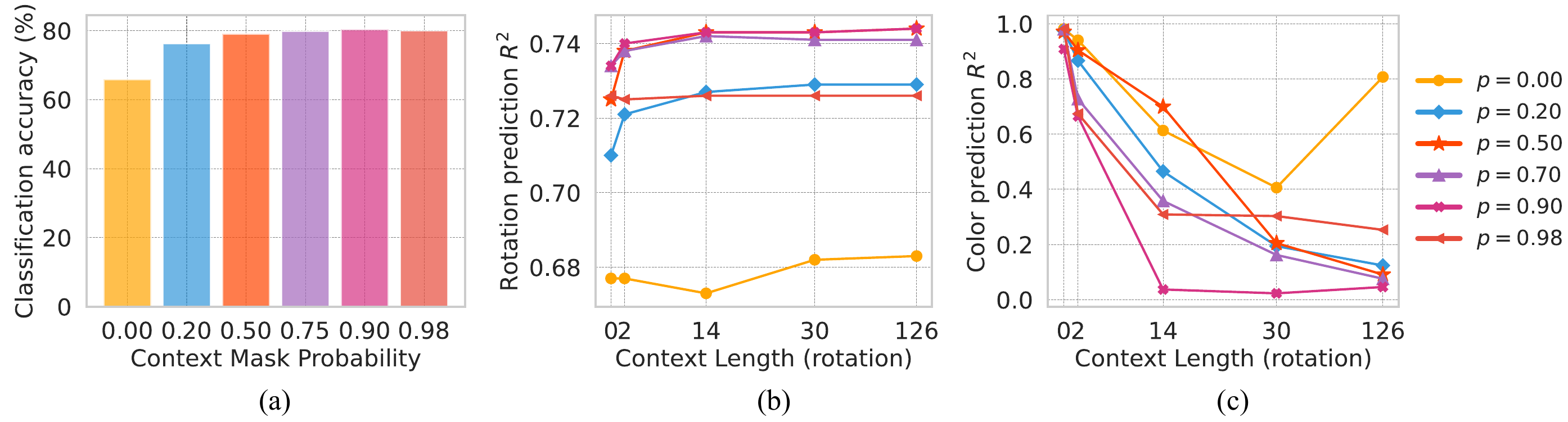}
    \caption{Role of context mask to avoid context-based shortcuts in \algo}
    \label{fig:fig_ablations_mask}
\end{figure}

\textbf{Role of Context Mask.} To illustrate how context masking effectively eliminates shortcuts, we conduct an ablation study with varying masking probabilities, detailed in~\Cref{fig:fig_ablations_mask}. We observed that as masking probability increases, performance on both classification and prediction tasks initially improves but later declines, reaching optimal performance at a masking probability of 90\%. Specifically, without the context mask, the trained model shows poor classification accuracy and weak performance on equivariant tasks. However, with just 20\% masking, performance significantly improves for both invariant and equivariant tasks.

\begin{figure}[!htb]
\centering
\begin{minipage}{0.35\textwidth}
\centering
\includegraphics[width=\textwidth]{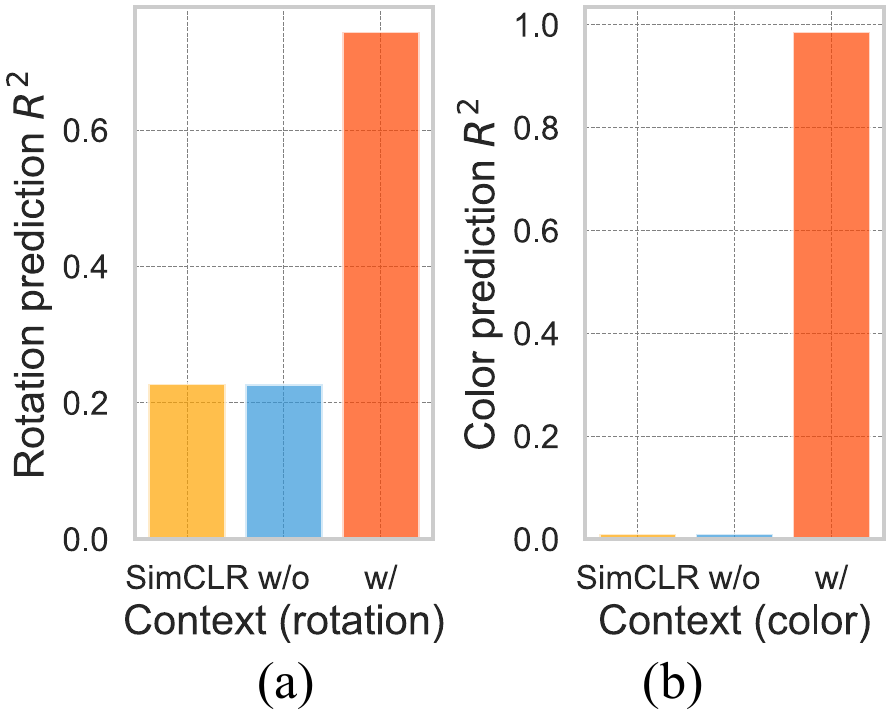}
\caption{Role of auxiliary predictor to avoid collapse to the trivial solution of invariance.}
\label{fig:ablation_aug_pred}
\end{minipage}
\hfill
\begin{minipage}{0.6\textwidth}
\centering
\includegraphics[width=\textwidth]{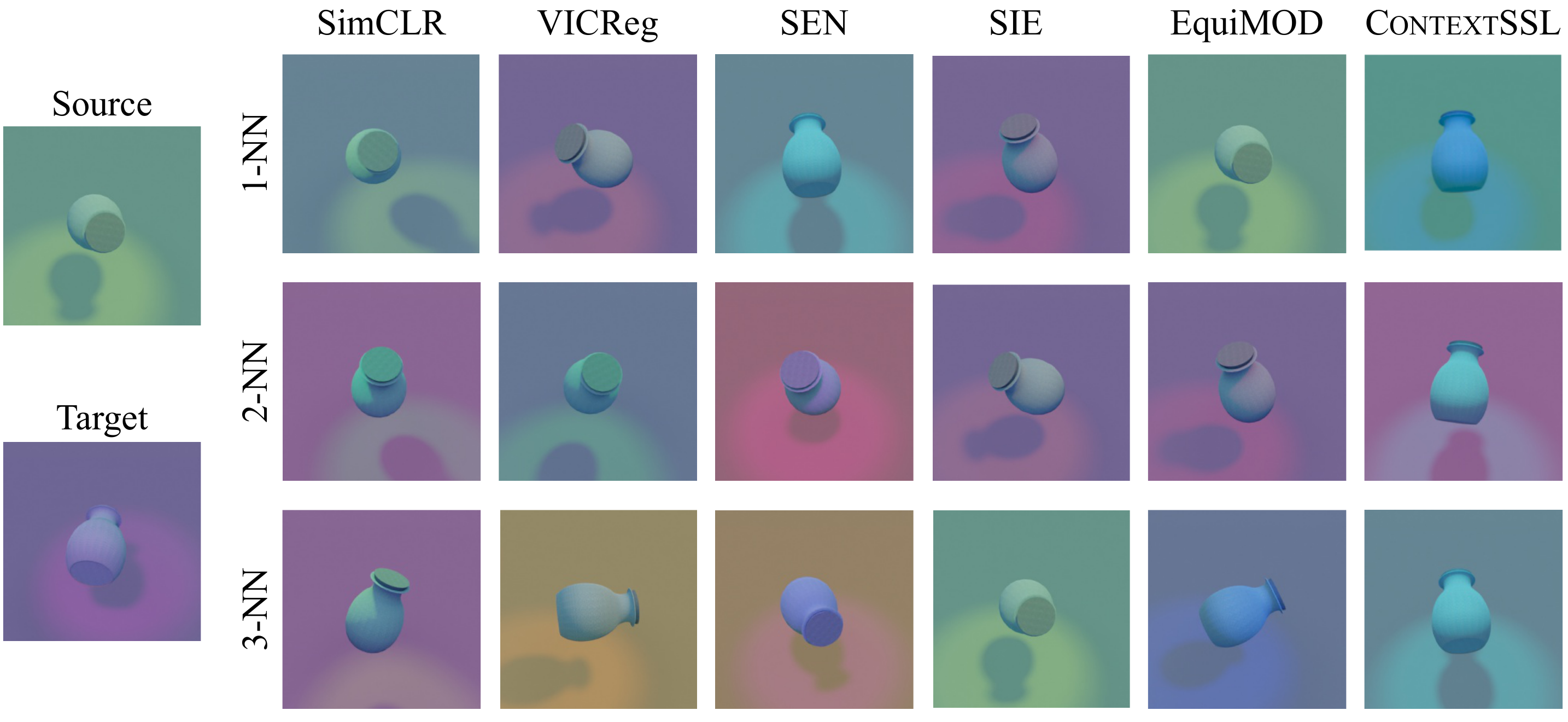}
\caption{Nearest neighbors of different methods taking as input the source image
and rotation angle. \algo aligns best with the rotation angle of the target image.}
    \label{fig:knn_baselines}
\end{minipage}
\end{figure}

\textbf{Role of Auxiliary Predictor.} We demonstrate that the auxiliary predictor is crucial for the model to achieve equivariance. In its absence, as depicted in \Cref{fig:ablation_aug_pred}, while the model retains its performance on the invariant classification task, it fails to learn equivariance and cannot effectively adapt to different contexts.

\subsection{Qualitative Assessment of Adaptation to Task-Specific Symmetries}

We conduct a qualitative assessment of model performance by taking the nearest neighbors of the predictor output when inputting a source image and a transformation variable, as shown in \Cref{fig:knn_baselines}. The nearest neighbors of invariance models (SimCLR and VICReg) have random rotation angles. Equivariance baselines (SEN, SIE, EquiMOD) correctly generate the target rotation angle for some of the three nearest neighbors but fail in others. \algo consistently outperforms all baseline models by accurately identifying the target angle in all instances of the top three nearest neighbors while remaining invariant to color variations. Additional qualitative assessments of \algo across varying contexts are provided in Section~\Cref{sec:knn_supplement}. Specifically, we demonstrate how the nearest neighbors for a source image vary with changes in context length, further emphasizing \algo's adaptability to become more invariant or equivariant depending on the context.

\subsection{Expanding to Diverse Data Transformations}
Previously, our experiments with the 3DIEBench dataset focused on rotations and color as transformation groups.
We extend our approach to transformations such as blurring, color jitter, and cropping on CIFAR10. 
We evaluate our approach using contexts derived from two augmentations simultaneously. 
The results for the combinations of crop and blur are reported in~\Cref{table:cifar10_results_cropblur}. Consistent with our previous results, while almost retaining the classification performance as SimCLR, \algo learns to adaptively enforce equivariance to crop (blur) and invariance to blur (crop) depending on context. Note that the invariance performance initially improves with increasing context length but then diminishes. This occurs due to the 90\% random masking ratio during training, which necessitates out-of-distribution generalization when the context length is large. Results on additional transformation pairs are provided in~\Cref{sec:supplement_cifar10_all}.

\begin{table*}[!htb]
    \centering
    \caption{Performance of \algo on invariant (classification) and equivariant (crop prediction, blur prediction) tasks in CIFAR-10 under the environment of crop, i.e. \algo (crop), and blur, i.e. \algo (blur).}
    \resizebox{1.\textwidth}{!}{%
    \begin{tabular}{lccccccccccc}
        \toprule
        Method & \multicolumn{5}{c}{Crop prediction ($R^2$)} & \multicolumn{5}{c}{Blur prediction ($R^2$)} & \multicolumn{1}{c}{Classification (top-1)}  \\
        \cmidrule(lr){2-6} \cmidrule(lr){7-11}\cmidrule(lr){12-12}
        & 0 & 2 & 14 & 30 & 126 & 0 & 2 & 14 & 30 & 126 & Representation \\
        \midrule
        SimCLR  & \multicolumn{5}{c}{0.459} & \multicolumn{5}{c}{0.371} & 89.1 \\
        SimCLR$^+$ (c=0)& \multicolumn{5}{c}{0.448} & \multicolumn{5}{c}{0.361} & 88.9 \\
        SimCLR$^+$ & \multicolumn{5}{c}{0.505} & \multicolumn{5}{c}{0.381} & 89.7 \\
        \algo (crop) & 0.608 & 0.607 & 0.607 & 0.608 & 0.608 & 0.920 & 0.854 & 0.624 & 0.667 & 0.694 & 88.5 \\
        \algo (blur) & 0.609 & 0.482 & 0.434 & 0.417 & 0.465 & 0.920 & 0.923 & 0.925 & 0.925 & 0.925 & 88.5 \\
        \bottomrule
    \end{tabular}
    }
       \label{table:cifar10_results_cropblur}
\end{table*}

\subsection{Enforcing Invariance or Equivariance to the Same Transformation Using Context}\label{sec:single env}
Apart from adaptively learning equivariance to a subset of transformation groups and invariance to the rest as shown in~\Cref{table:main_table}, we extend \algo to operate within environments characterized by a single transformation. Motivated by this, we ask the question: \emph{Can \algo adapt to learn equivariance or invariance to the same transformation depending on the context?} At training, we randomly sample one of these environments. If the environment corresponds to enforcing equivariance, we construct our context in the same way as before i.e. pairs of positives transformed using augmentations sampled from the transformation group. However, if the environment corresponds to enforcing invariance, we maximize alignment between positives transformed by augmentation sampled from the transformation group without conditioning on that augmentation. Take rotation in 3DIEBench as an example. As shown in~\Cref{table:main_single_environment}, and similar to findings in two transformation setting (rotation and color) in \Cref{table:main_table}, \algoname effectively adapts to enforce invariance and equivariance to rotation depending on the context. Specifically, under the 'none' context, where context is defined as $\{(x_i, 0, y_i)\}_{i=1}^K$, the model's performance on rotation prediction task improves with context Detailed results for predicting individual transformation parameters are provided in \Cref{table:single_environment_indi} in the Appendix.

\begin{table*}[!htb]
    \centering
    \caption{Performance of \algo on 3DIEBench to enforce equivariant and invariance to the same transformation depending on context. We observe that under the equivariant environment (rotation), \algo performs well on rotation prediction $R^2$ (equivariant task), and under the invariant environment (none), \algo performs poorly on the equivariant task, thus enforcing invariance to rotations.}
    \resizebox{.7\textwidth}{!}{%
    \begin{tabular}{lccccccc}
        \toprule
        Method & \multicolumn{5}{c}{Rotation prediction ($R^2$)}  & \multicolumn{1}{c}{Classification (top-1)}  \\
        \cmidrule(lr){2-6} \cmidrule(lr){7-7}
        & 0 & 2 & 14 & 30 & 126 & Representation \\
        \midrule
        SimCLR  & \multicolumn{5}{c}{0.506} & 85.3 \\
        SimCLR$^+$ (c=0)& \multicolumn{5}{c}{0.478} & 83.4 \\
        SimCLR$^+$ & \multicolumn{5}{c}{0.489} & 81.0 \\
        \algo (rotation) & 0.737 & 0.737 & 0.736 & 0.737 & 0.738 & 80.6 \\
        \algo (none) & 0.737 & 0.717 & 0.477 & 0.377 & 0.473 & 80.6 \\
        \bottomrule
    \end{tabular}
    }
       \label{table:main_single_environment}
\end{table*}


\subsection{Context World Models Beyond Self-Supervised Learning}

While our analysis has primarily focused on self-supervised learning, the concept of context is versatile and extends beyond representation learning. In principle, irrespective of the task at hand, paying attention to context can learn and identify features defined by it. To validate this and explore broader applications of our algorithm, we consider a supervised learning task where our transformer model is trained to directly predict the labels corresponding to an input image. We further corrupt the labels to be directly influenced by the augmentation group transforming the data. Specifically, for the 3DIEBench dataset, we add a constant value of 10 to each label if the context corresponds to the rotation group and leave it unchanged otherwise. Note that, since the shortcuts caused by positive and negative samples in self-supervised learning are absent in the supervised setting, the context mask is not applied. We report classification performance along with rotation and color prediction equivariant measures. As shown in~\Cref{table:main_supervised_table}, \algo's classification accuracy improves with context, demonstrating its ability to better identify the underlying symmetry group with an increase in context. Additional results are provided in~\Cref{sec:supervised_supplement}. Further, \algo serves as a general framework that can adapt to different training regimes, such as supervised learning. 

\begin{table*}[!htb]
    \centering
    \caption{Performance of \algo on equivariant  tasks (including classificaion) for context-dependent labels. \algo adapts to context-dependent labels with varying context.}
    \resizebox{1.\textwidth}{!}{%
    \begin{tabular}{lccccccccccccccc}
        \toprule
        Method & \multicolumn{5}{c}{Rotation prediction ($R^2$)} & \multicolumn{5}{c}{Color prediction ($R^2$)} & \multicolumn{5}{c}{Classification (top-1)}  \\
        \cmidrule(lr){2-6} \cmidrule(lr){7-11}\cmidrule(lr){12-16}
        & 0 & 2 & 14 & 30 & 126 & 0 & 2 & 14 & 30 & 126 & 0 & 2 & 14 & 30 & 126 \\
        \midrule
        SimCLR (color) & \multicolumn{5}{c}{0.537} & \multicolumn{5}{c}{0.056} & \multicolumn{5}{c}{72.0} \\
        SimCLR (rotation) & \multicolumn{5}{c}{0.537} & \multicolumn{5}{c}{0.056} &\multicolumn{5}{c}{14.2} \\
        SimCLR$^+$ (c=0) (color) & \multicolumn{5}{c}{0.427} & \multicolumn{5}{c}{-0.007} & \multicolumn{5}{c}{80.4}\\
        SimCLR$^+$ (c=0) (rotation) & \multicolumn{5}{c}{0.427} & \multicolumn{5}{c}{-0.007} & \multicolumn{5}{c}{5.2} \\
        SimCLR$^+$ (color) & \multicolumn{5}{c}{0.424} & \multicolumn{5}{c}{0.243} & 16.8 & 15.1 & 15.6 & 14.8 & 14.0  \\
        SimCLR$^+$ (rotation) & \multicolumn{5}{c}{0.424} & \multicolumn{5}{c}{0.243} & 56.1 & 58.2 & 58.4 & 58.4 & 59.1  \\
        \algo (color) & 0.556 & 0.542 & 0.538 & 0.540 & 0.539 & 0.913 & 0.973 & 0.981 & 0.982 & 0.982 & 8.9 & 82.4 & 82.7 & 82.8 & 83.0 \\
        \algo (rotation) & 0.556 & 0.624 & 0.661 & 0.665 & 0.666 & 0.913 & 0.379 & 0.111 & 0.095 & 0.093 & 73.5 & 82.7 & 82.6 & 82.6 & 83.0 \\
        \bottomrule
    \end{tabular}
    }
       \label{table:main_supervised_table}
\end{table*}


%% file: related.tex
\section{Related Work}

\textbf{Self-Supervised Learning.} 
Existing SSL methods generally belong to two categories: invariant learning~\citep{simclr,bardes2021vicreg,chen2021exploring,he2020momentum,zbontar2021barlow,grill2020bootstrap} and equivariant learning. The representative method for invariant learning is contrastive learning, which draws the representations of positive samples together in the latent space such that the representations are invariant to data augmentation. 
Contrastive learning can learn highly discriminative features at the cost of losing certain image information due to the invariance constraint \cite{xiao2020should}. Motivated by this limitation, recent works explore merging contrastive learning with equivariant learning tasks by separate embedding \cite{xiao2020should,sie}, augmentation-conditioned predictor \cite{devillers2022equimod,iwm}, and explicit equivariant transformation \cite{gupta2023structuring}. However, existing works still inherit the limitations of contrastive learning: its symmetry prior is built on a given set of manual augmentations and is not adaptive to downstream tasks. In contrast, our method enables the contextual world model to adapt its symmetry to the contextual data, which is more flexible and generalizable to various tasks.

\textbf{World Models.} 
World modeling has achieved notable success in reinforcement learning (RL) for model-based planning~\cite{ha2018world,sekar2020planning,hafner2019dream} and vision~\citep{hafner2023mastering,hu2023gaia,yang2023learning}, where it involves predicting future states based on current observations and actions. This concept, however, has not yet been fully leveraged in visual representation learning. Nevertheless, ~\citet{iwm} shows that several families of self-supervised learning approaches can be reformulated through the lens of world modeling.
Equivariant self-supervised learning methods. Specifically, Masked Image Modeling approaches~\citep{mae,bao2021beit,el2024scalable,xie2022simmim} consider masked pixels and target pixel reconstruction as their action and next state. Other equivariant learning approaches~\citep{devillers2022equimod,park2022learning,sie} consider data transformations and representation of the target image as their action and next state pair. However, unlike true world modeling, these approaches do not track past experiences, a component critical for generalization. Our method instead leverages context to track past experiences in terms of state, action, and next-state triplets, enabling it to adapt and generalize to varying environments.

\textbf{In-context Learning.} Our work is inspired by and extends the concept of in-context learning (ICL)~\citep{brown2020language} to training. Initially studied in the context of language, in-context learning has recently been adapted for vision tasks~\citep{gupta2023context,wang2023images,bar2022visual,li2021prefix}, allowing models to infer environmental features or tasks directly from input prompts without predefined notions. For example, Visual Prompting~\citep{wang2023images,bar2022visual} uses a task input/output example pair and a query image at test time, and uses inpainting to generate the desired output. ~\citet{gupta2023context} propose using unlabeled data as context at training to extract environment-specific signals and address domain generalization. ICL has been extensively explored in various domains, including vision, language, and multimodal tasks. However, our work is the first to apply ICL to vision self-supervised representation learning.


%% file: conclusion.tex
\section{Conclusion and Future Perspectives}
\label{sec:conclusion}
The field of language modeling has witnessed a significant paradigm shift over the past decade, moving towards foundation models that generalize across a variety of tasks either directly or through distillation. However, this shift toward generalization has been conspicuously absent in the vision domain. This is largely because self-supervised approaches for vision still heavily rely on inductive priors strongly introduced by enforcing either invariance or equivariance to data augmentations. This renders representations brittle in downstream tasks that do not conform to these priors and necessitates retraining the representation separately for each task. This work forgoes any notion of pre-defined symmetries and instead trains a model to infer the task-relevant symmetries directly from the context through what we term Contextual Self-Supervised Learning (\algo). The ability of our model to learn selective equivariances and invariances based on mere context opens up new avenues for effectively handling a broader range of tasks, particularly in dynamic environments where the relevance of specific features may change over time. 
However, we limit our scope of symmetries to hand-crafted transformations in the data and do not explore naturally occurring symmetries. 
%
Nonetheless, \algo lays the groundwork for models that can potentially discern and adapt to the underlying patterns of tasks, recognize shortcuts, and more effectively generalize across unseen scenarios. Through this work, we hope to contribute to a broader understanding of how machines can learn more like humans --- contextually, adaptively, and with an eye toward the infinite variability of the real world.

%% file: acknowledgement.tex
\section{Acknowledgement}
This research was supported in part by
Office of Naval Research grant N00014-20-1-2023 (MURI ML-SCOPE), NSF AI Institute TILOS (NSF CCF-2112665), and the Alexander von Humboldt Foundation. CW and TJ acknowledge support from NSF Expeditions grant (award 1918839: Collaborative Research: Understanding the World Through Code) and Machine Learning for Pharmaceutical Discovery and Synthesis (MLPDS) consortium. We acknowledge MIT SuperCloud and Lincoln Laboratory Supercomputing Center for providing HPC resources that have contributed to this work.

%% file: appendix.tex
\newpage
\appendix
\startcontents[appendices]
\printcontents[appendices]{}{1}{\section*{Appendix}}

\section{Supplementary experimental details and assets disclosure}\label{sec:exp_setup}
To evaluate the efficacy of our proposed algorithm \algoname, our experiments are designed to address the following questions:
\begin{enumerate}
    \item How does \algoname fare against competitive invariant and equivariant self-supervised learning approaches in terms of performance across varying context sizes and different sets of data transformations?
    \item How effectively can \algoname identify task-specific symmetries, both within the scope of self-supervised learning and beyond?
    \item What roles do specific components such as selective masking and the auxiliary latent transformation predictor play in facilitating the learning of general and context-adaptable representations?
\end{enumerate}

\subsection{Assets}
We do not introduce new data in the course of this work. Instead, we use publicly available widely used image datasets for the purposes of benchmarking and comparison.

\subsection{Hardware and setup}
\label{sec.compute}
Each experiment was conducted on 1 NVIDIA Tesla V100 GPUs, each with 32GB of accelerator RAM. The CPUs used were Intel Xeon E5-2698 v4 processors with 20 cores and 384GB of RAM. All experiments were implemented using the PyTorch deep learning framework.

\subsection{Datasets}\label{sec:datasets}
\textbf{3D Invariant Equivariant Benchmark (3DIEBench).} To test equivariance and invariance to multiple data transformations, we use the 3D Invariant Equivariant Benchmark (3DIEBench)~\citep{sie} which has been specifically designed to address the limitations of existing datasets in evaluating invariant and equivariant representations. It contains images of 3D objects along with their latent parameters such as object rotation, lighting color, and floor color. Since we have access available to individual meta latent parameters, transformation parameters between two views of an object are calculated as the difference between their individual latents. We test our approach on 3DIEBench under two settings 1) Considering two transformation groups: rotation and color with the aim of learning invariance to one and equivariance to another after conditioning on context; 2) Considering one transformation group, say rotation and learning to enforce invariance or equivariance to rotation with context. As previously mentioned, all methods are trained for 1000 epochs using a batch size of 512 on 128$\times$128 resolution images. We use the standard training, validation and test splits, made publicly available by the authors~\citep{sie}.
\\

\textbf{CIFAR10.} 3DIEBench dataset is limited to only rotations and color as transformation groups. We extend our approach to include more common self-supervised benchmarks, such as CIFAR-10, incorporating transformations like blurring, color jitter, and cropping. Unlike 3DIEBench, we manually construct positive pairs by applying compositions of these handcrafted augmentations. We consider three transformation groups: crop, blur and color. Similar to 3DIEBench, we consider combinations of two groups for each training run. We use the standard training, validation and test splits.

\subsection{Baseline Algorithms} Among the invariant self-supervised approached, we compare our approach to VICReg~\citep{bardes2021vicreg} and and SimCLR~\citep{simclr}. For each method, comparisons are drawn using their originally proposed architectures. For the equivariant baselines, we consider EquiMOD~\citep{devillers2022equimod}, SIE~\citep{sie} and SEN~\citep{park2022learning}. Similar to~\citet{sie}, For SEN, we use the InfoNCE loss instead the original triplet loss. To discard the performance gains potentially arising from \algoname's transformer architecture, for each approach, we consider an additional baseline that replaces the original projection heads or predictor with our transformer model. Given an algorithm name $\mathcal{N}$, we refer to this baseline as $\mathcal{N}^+$. Amongst these, we report the best performing variant in our results. For $\mathcal{N}^+$, we conduct analysis in two distinct settings: 1) a 'no context' or $c=0$ invariant condition, and 2) a fully contextualized setting with a context length of 126.

\subsection{Training Protocol}
To ensure a fair comparison across different algorithms for each dataset, we use a standardized neural
network backbone. Precisely, for our encoder, we use a ResNet-18 backbone pre-trained on ImageNet. For \algo, output features from the encoder are transformed into the context sequence, which is then processed by the decoder-only Transformer~\citep{vaswani2017attention} from the GPT-2 Transformer family~\citep{radford2019language}. Our model configuration includes 3 layers, 4 attention heads, and a 2048-dimensional embedding space, consistently applied across all datasets. Linear layers are utilized to convert the input sequence into the transformer's latent embedding of dimension 2048 and to map the predicted output vectors to the output space of dimension 512. 

We fix the maximum training context length to 128. Since for every $y$, the corresponding token ($x_i, a_i$) is masked out, context length $L$ corresponds to effective context length $L-2$. Thus, we report \algo's performance over varying test context length of 0, 2, 14, 30 and 126. On all datasets, we train \algo with the Adam optimizer with a learning rate of $5e^{-5}$ and weight decay $1e{-3}$. For baseline self-supervised approaches, in their original architecture, we use a learning rate of $1e^{-3}$ with no weight decay. However, when tested using the transformer architecture, we choose one of the above two optimizer hyperpameters. Consequently, performance of the best performing model is reported among the two baselines. Similar to~\citet{sie}, we report hyper-parameters and architectures specific to each method:
\begin{itemize}
    \item \textbf{SimCLR~\citep{simclr}} We train using a 2048-2048-2048 dimensional multi-layered perceptron (MLP) based projection head with a temperature of 0.5.
    \item \textbf{VICReg~\citep{bardes2021vicreg}} We train using a 2048-2048-2048 MLP for the projection head and use weight of 10 for both the invariance loss and variance loss and 1 for covariance loss.  
    \item \textbf{SEN~\citep{park2022learning}} Similar to other approaches we use a projection head of dimension 2048-2048-2048 and temperature 0.1. 
    \item \textbf{EquiMod~\citep{devillers2022equimod}} We use the standatd projection head of dimensions 1024-1024-128 and use equal weighing of the invariance and the equivariance loss.
    \item \textbf{SIE~\citep{sie}} We use two 1024-1024-1024 projection heads, one for invariant latent space and other for equivariant. When trained to learn equivariance to only rotation or only color, we use weight of 10 for both the invariance loss and variance loss, 1 for the covariance loss and 4.5 for the equivariant loss. However, when trained to be equivariant to both rotation and color jointly, we use 10 as the equivariant weight. 
\end{itemize}

\subsection{Evaluation metrics}
In line with established self-supervised learning methodologies, we begin by assessing the quality of the learned representations through downstream tasks. For evaluating invariant representations, we employ linear classification over frozen features. To evaluate equivariant representations, we predict the corresponding data transformation. This prediction takes representations from two differently transformed views of the same object and regresses on the applied transformation between them.  Further, we use Mean Reciprocal Rank (MRR) and Hit Rate at $k$ (H@k) to evaluate the performance for our context predictor. Given the source data and the transformation action, we identify the $k$ nearest neighbors in the embedding space. MRR is calculated as the average reciprocal rank of the target embedding within these nearest neighbors. Hit rate-k (H@k) assigns a score of 1 if the target embedding is within the 
k-nearest neighbors of the predicted embedding and 0 otherwise. Similar to ~\citet{sie}, we restrict the search for nearest neighbors to different views of the same object, thus ensuring that the predictor is not penalized for retrieving an incorrect object in a pose similar to the correct one.

\section{Additional Experiments}
\subsection{Quantitative Assessment of Adaptation to Task-Specific Symmetries}
\label{sec.main.suppl}

In this section, we present additional results on the quantitative assessment of model performance on 3DIEBench, including the evaluation of learned representations on equivariant tasks (rotation and color prediction) to predict individual latent values. In contrast, the results in \Cref{table:main_table} focus on predicting relative latent values between pairs of image embeddings as inputs.

\begin{table*}[!htb]
    \centering
    \caption{Quantitative evaluation of learned representations on equivariant (rotation prediction, color prediction) tasks to predict individual latent values.}
    \resizebox{1.\textwidth}{!}{%
    \begin{tabular}{llccccccccccc}
        \toprule
        $\gG$ & Method & \multicolumn{5}{c}{Rotation prediction ($R^2$)} & &\multicolumn{5}{c}{Color prediction ($R^2$)}  \\
        \cmidrule(lr){3-7} \cmidrule(lr){9-13}
        & & 0 & 2 & 14 & 30 & 126 & & 0 & 2 & 14 & 30 & 126 \\
        \midrule
        & \multicolumn{1}{l}{\textit{\textcolor{gray}{Invariant }}}\\
        & SimCLR & \multicolumn{5}{c}{0.791} & & \multicolumn{5}{c}{0.137} \\
        & SimCLR$^+$(c=0) & \multicolumn{5}{c}{0.773} & &\multicolumn{5}{c}{0.061} \\
        & SimCLR$^+$ & \multicolumn{5}{c}{0.773} & &\multicolumn{5}{c}{0.116}  \\
        & VICReg & \multicolumn{5}{c}{0.660} & & \multicolumn{5}{c}{0.011} \\
        & VICReg$^+$(c=0) & \multicolumn{5}{c}{0.615} & & \multicolumn{5}{c}{0.061} \\
        \midrule
        & \multicolumn{1}{l}{\textit{\textcolor{gray}{Equivariant }}}\\
        \multirow{3}{*}{\rotatebox[origin=c]{90}{\parbox{0.8cm}{Rotation \\ + Color}}} & EquiMOD & \multicolumn{5}{c}{0.712} && \multicolumn{5}{c}{0.221} \\
        & SIE & \multicolumn{5}{c}{\textbf{0.760}} && \multicolumn{5}{c}{\textbf{0.972}} \\
        & SEN & \multicolumn{5}{c}{0.617} && \multicolumn{5}{c}{0.888} \\
        \midrule
        \multirow{4}{*}{\rotatebox[origin=c]{90}{Rotation}} & EquiMOD & \multicolumn{5}{c}{0.707} && \multicolumn{5}{c}{0.033} \\
        & SIE & \multicolumn{5}{c}{0.790} && \multicolumn{5}{c}{\textbf{0.001}} \\
        & SEN & \multicolumn{5}{c}{0.723} && \multicolumn{5}{c}{0.437} \\
        & \algo\footnotemark & 0.838 & 0.839 & 0.840 & 0.840 & \textbf{0.840} & & 0.895 & 0.620 & 0.021 & 0.014 & 0.021  \\
        \midrule
        \multirow{4}{*}{\rotatebox[origin=c]{90}{Color}} & EquiMOD & \multicolumn{5}{c}{0.660} && \multicolumn{5}{c}{0.855} \\
        & SIE & \multicolumn{5}{c}{\textbf{0.560}} && \multicolumn{5}{c}{0.974} \\
        & SEN & \multicolumn{5}{c}{0.713} && \multicolumn{5}{c}{0.876} \\
        & \algo\footnotemark & 0.838 & 0.800 & 0.699 & 0.666 & 0.685 && 0.895 & 0.981 & 0.985 & 0.985 & \textbf{0.986}  \\        
        \bottomrule
    \end{tabular}
    }
       \label{table:main_table_individual_version}
\end{table*}

\subsubsection{Invariant Classification and Equivariant transformation prediction task}

As shown in~\Cref{table:main_table_individual_version}, invariant self-supervised learning methods such as SimCLR and VICReg underperform in equivariant augmentation prediction tasks. The equivariant baselines, EquiMOD, SIE, and SEN, exhibit improvements compared to the invariant baselines in some of the augmentation prediction tasks. 
However, their degree of equivariance is much worse compared to \algo. Besides, aligning them with different targeted symmetry groups requires retraining the entire model. In contrast, \algo employs a single model capable of learning equivariance to rotation and invariance to color (or vice versa) based on the given context.
As seen from the two rows corresponding to \algo~\Cref{table:main_table}, when the context corresponds to pairs of data with transformations sampled from the rotation (color) group, the model adaptively learns to be invariant to color (rotation) while retaining equivariance to rotation (color).

Results in \Cref{table:main_table} are the average value over three random seeds. We provide the standard deviation for rotation and color prediction of \algo in \Cref{table:ours_rot_stddev} and \Cref{table:ours_color_stddev}.

\begin{table*}[!htb]
    \centering
    \caption{Performance of \algo in 3DIEBench in rotation prediction under the environment of rotation, i.e. \algo (rotation), and color, i.e. \algo (color), with standard deviations over three random seeds.}
    \resizebox{1.\textwidth}{!}{%
    \begin{tabular}{lcccccc}
        \toprule
        Method & \multicolumn{5}{c}{Rotation prediction ($R^2$)}  \\
        \cmidrule(lr){2-6}
        & 0 & 2 & 14 & 30 & 126 \\
        \midrule
        \algo (rotation) & 0.734 $\pm$ 0.002 & 0.740 $\pm$ 0.004 & 0.743 $\pm$ 0.001 & 0.743 $\pm$ 0.001 & 0.744 $\pm$ 0.001 \\
        \algo (color) & 0.735 $\pm$ 0.001 & 0.614 $\pm$ 0.108 & 0.389 $\pm$ 0.054 & 0.345 $\pm$ 0.040 & 0.344 $\pm$ 0.003 \\
        \bottomrule
    \end{tabular}
    }
       \label{table:ours_rot_stddev}
\end{table*}

\begin{table*}[!htb]
    \centering
    \caption{Performance of \algo in 3DIEBench in color prediction under the environment of rotation, i.e. \algo (rotation), and color, i.e. \algo (color), with standard deviations over three random seeds.}
    \resizebox{1.\textwidth}{!}{%
    \begin{tabular}{lcccccc}
        \toprule
        Method & \multicolumn{5}{c}{Color prediction ($R^2$)}  \\
        \cmidrule(lr){2-6}
        & 0 & 2 & 14 & 30 & 126 \\
        \midrule
        \algo (rotation) & 0.908 $\pm$ 0.002 & 0.664 $\pm$ 0.166 & 0.037 $\pm$ 0.010 & 0.023 $\pm$ 0.001 & 0.046 $\pm$ 0.007 \\
        \algo (color) & 0.908 $\pm$ 0.002 & 0.981 $\pm$ 0.002 & 0.985 $\pm$ 0.001 & 0.986 $\pm$ 0.001 & 0.986 $\pm$ 0.001 \\
        \bottomrule
    \end{tabular}
    }
       \label{table:ours_color_stddev}
\end{table*}

\subsubsection{Equivariant Measures Based on Nearest Neighbours Retrieval}
Similar to \Cref{table:mrr}, we provide the performance of \algo on MRR and H@k compared to baseline methods with trained equivariance to rotation. While \Cref{table:mrr} uses the validation set data as the retrieval library, \Cref{table:mrr_train} provides the results using the training set data.
\algo outperforms the baseline models, and its performance on all the metrics consistently improves with increasing context length, showing adaptation to rotation-specific features.

\begin{table*}[!htb]
    \centering
    \caption{Quantitative evaluation of learned predictors equivariant to only rotation based on Mean Reciprocal Rank (MRR) and Hit Rate H@k on training dataset. \algo learns to be more equivariant to rotation with context.}
    \resizebox{1.\textwidth}{!}{%
    \begin{tabular}{lccccccccccccccccc}
        \toprule
        Method & \multicolumn{5}{c}{MRR ($\uparrow$)} && \multicolumn{5}{c}{H@1 ($\uparrow$)} && \multicolumn{5}{c}{H@5 ($\uparrow$)} \\
        \cmidrule(lr){2-6} \cmidrule(lr){8-12}\cmidrule(lr){14-18}
        & 0 & 2 & 14 & 30 & 126 && 0 & 2 & 14 & 30 & 126 && 0 & 2 & 14 & 30 & 126 \\ 
        \midrule
        EquiMOD & \multicolumn{5}{c}{0.17} && \multicolumn{5}{c}{0.06} && \multicolumn{5}{c}{0.24} \\
        SEN & \multicolumn{5}{c}{0.17} && \multicolumn{5}{c}{0.06} && \multicolumn{5}{c}{0.24} \\
        \algo & 0.282 & 0.321 & 0.470 & 0.498 & \textbf{0.531} && 0.132 & 0.263 & 0.375 & 0.398 & \textbf{0.402} && 0.436 & 0.495 & 0.650 & 0.669 & \textbf{0.680}\\
        \bottomrule
    \end{tabular}
    }
\label{table:mrr_train}
\end{table*}

\subsection{Role of Context Mask and Auxiliary Predictor}
\label{sec.mask.suppl}
In this section, we provide additional results for the role of context mask and auxiliary predictor.

\subsubsection{Role of Context Mask}

\begin{figure}[!htb]
    \centering
    \includegraphics[width=\textwidth]{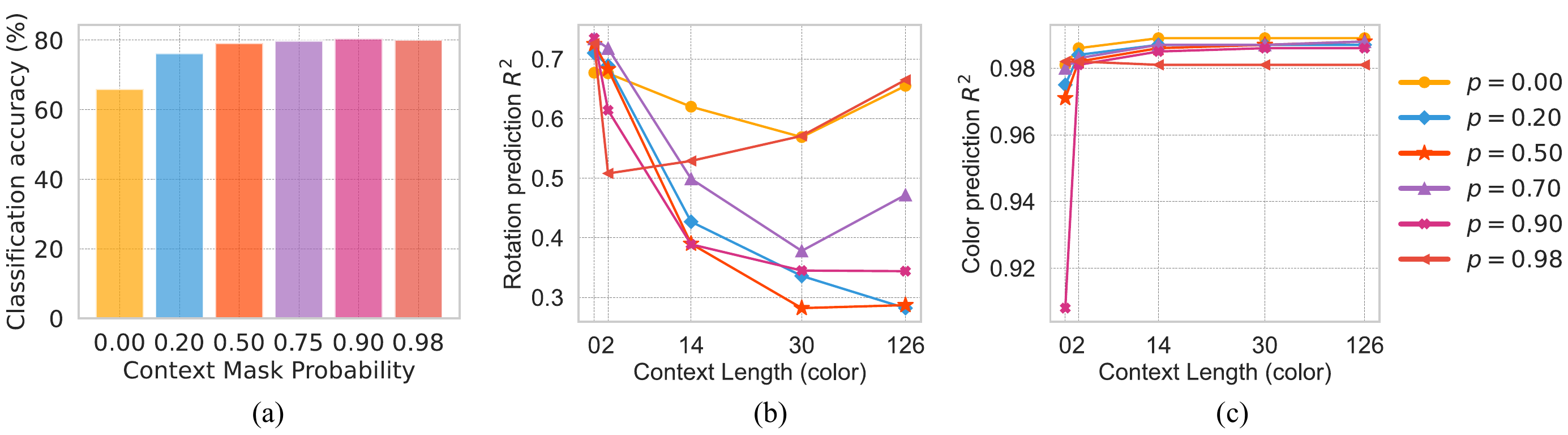}
    \caption{Role of context mask to avoid context based shortcuts in \algo under color context}
    \label{fig:fig_ablations_mask_supplement}
\end{figure}

In addition to \Cref{fig:fig_ablations_mask}, we provide the performance of the rotation and color prediction tasks with varying masking probabilities under the environment of color in \Cref{fig:fig_ablations_mask_supplement}. We observe that as masking probability increases, performance on both classification and prediction tasks initially improves but later declines, reaching optimal performance at a masking probability of 90\%.

Results in \Cref{fig:fig_ablations_mask} and \Cref{fig:fig_ablations_mask_supplement} are the average value over three random seeds. We provide the standard deviation for rotation and color prediction of \algo in \Cref{table:masking_rot_std} and \Cref{table:masking_color_std}.

\begin{table*}[!htb]
    \centering
    \caption{Performance of \algo rotation prediction tasks in 3DIEBench under different random masking probabilities, with standard deviations over three random seeds.}
    \resizebox{.9\textwidth}{!}{%
    \begin{tabular}{lcccccc}
        \toprule
        Context & Probability & \multicolumn{5}{c}{Rotation prediction ($R^2$)} \\
        \cmidrule(lr){3-7} 
        & & 0 & 2 & 14 & 30 & 126 \\
        \midrule
        \multirow{6}{*}{Rotation} & 0.00 & 0.677 $\pm$ 0.004 & 0.677 $\pm$ 0.002 & 0.673 
 $\pm$ 0.009 & 0.682 $\pm$ 0.003 & 0.683 $\pm$ 0.003 \\
        & 0.20 & 0.710 $\pm$ 0.002  & 0.721 $\pm$ 0.006 & 0.727 $\pm$ 0.002 & 0.729 $\pm$ 0.001 & 0.729 $\pm$ 0.001 \\
        & 0.50 & 0.725 $\pm$ 0.001 & 0.738 $\pm$ 0.005 & \textbf{0.743 $\pm$ 0.001} & \textbf{0.743 $\pm$ 0.001} & \textbf{0.744 $\pm$ 0.001} \\
        & 0.75 & \textbf{0.734 $\pm$ 0.002} & 0.738 $\pm$ 0.006 & 0.742 $\pm$ 0.004 & 0.741 $\pm$ 0.004 & 0.741 $\pm$ 0.002 \\
        & 0.90 & \textbf{0.734 $\pm$ 0.002} & \textbf{0.740 $\pm$ 0.004} & \textbf{0.743 $\pm$ 0.001} & \textbf{0.743 $\pm$ 0.001} & \textbf{0.744 $\pm$ 0.001} \\
        & 0.98 & 0.726 $\pm$ 0.002 & 0.725 $\pm$ 0.003 & 0.726 $\pm$ 0.002 & 0.726 $\pm$ 0.003 & 0.726 $\pm$ 0.003 \\
        \midrule
        \multirow{6}{*}{Color} & 0.00 & \textbf{0.677 $\pm$ 0.004} & 0.676 $\pm$ 0.005 & 0.620 
 $\pm$ 0.019 & 0.569 $\pm$ 0.019 & 0.655 $\pm$ 0.010 \\
        & 0.20 & 0.710 $\pm$ 0.002 & 0.689 $\pm$ 0.013 & 0.427 $\pm$ 0.031 & 0.336 $\pm$ 0.007 & \textbf{0.282 $\pm$ 0.022} \\
        & 0.50 & 0.725 $\pm$ 0.001 & 0.683 $\pm$ 0.006 & 0.390 $\pm$ 0.031 & \textbf{0.282 $\pm$ 0.013} & 0.287 $\pm$ 0.002 \\
        & 0.75 & 0.734 $\pm$ 0.002 & 0.718 $\pm$ 0.002 & 0.499 $\pm$ 0.035 & 0.378 $\pm$ 0.054 & 0.472 $\pm$ 0.015 \\
        & 0.90 & 0.735 $\pm$ 0.001 & 0.614 $\pm$ 0.108 & \textbf{0.389 $\pm$ 0.054} & 0.345 $\pm$ 0.040 & 0.344 $\pm$ 0.003 \\
        & 0.98 & 0.726 $\pm$ 0.002 & \textbf{0.508 $\pm$ 0.127} & 0.529 $\pm$ 0.141 & 0.571 $\pm$ 0.125 & 0.665 $\pm$ 0.023 \\
        \bottomrule
    \end{tabular}
    }
       \label{table:masking_rot_std}
\end{table*}

\begin{table*}[!htb]
    \centering
    \caption{Performance of \algo color prediction tasks in 3DIEBench under different random masking probabilities, with standard deviations over three random seeds.}
    \resizebox{.9\textwidth}{!}{%
    \begin{tabular}{lcccccc}
        \toprule
        Context & Probability & \multicolumn{5}{c}{Color prediction ($R^2$)} \\
        \cmidrule(lr){3-7}
        & & 0 & 2 & 14 & 30 & 126 \\
        \midrule
        \multirow{6}{*}{Rotation} & 0.00 & 0.981 $\pm$ 0.002 & 0.940 $\pm$ 0.033 & 0.613 $\pm$ 0.123 & 0.406 $\pm$ 0.125 & 0.807 $\pm$ 0.080 \\
        & 0.20 & 0.975 $\pm$ 0.001 & 0.866 $\pm$ 0.171 & 0.465 $\pm$ 0.113 & 0.194 $\pm$ 0.057 & 0.124 $\pm$ 0.027 \\
        & 0.50 & 0.971 $\pm$ 0.002 & 0.904 $\pm$ 0.086 & 0.699 $\pm$ 0.028 & 0.205 $\pm$ 0.054 & 0.091 $\pm$ 0.016 \\
        & 0.75 & 0.980 $\pm$ 0.001 & 0.727 $\pm$ 0.351 & 0.358 $\pm$ 0.233 & 0.162 $\pm$ 0.021 & 0.076 $\pm$ 0.009 \\
        & 0.90 & \textbf{0.908 $\pm$ 0.002} & \textbf{0.664 $\pm$ 0.166} & \textbf{0.037 $\pm$ 0.010} & \textbf{0.023 $\pm$ 0.001} & \textbf{0.046 $\pm$ 0.007} \\
        & 0.98 & 0.982 $\pm$ 0.001 & 0.674 $\pm$ 0.368 & 0.309 $\pm$ 0.139 & 0.303 $\pm$ 0.118 & 0.253 $\pm$ 0.033 \\
        \midrule
        \multirow{6}{*}{Color} & 0.00 & 0.981 $\pm$ 0.002 & \textbf{0.986 $\pm$ 0.002} & \textbf{0.989 $\pm$ 0.001} & \textbf{0.989 $\pm$ 0.001} & \textbf{0.989 $\pm$ 0.001} \\
        & 0.20 & 0.975 $\pm$ 0.001 & 0.984 $\pm$ 0.002 & 0.987 $\pm$ 0.001 & 0.987 $\pm$ 0.001 & 0.987 $\pm$ 0.001 \\
        & 0.50 & 0.971 $\pm$ 0.002 & 0.982 $\pm$ 0.002 & 0.986 $\pm$ 0.002 & 0.987 $\pm$ 0.002 & 0.988 $\pm$ 0.001 \\
        & 0.75 & 0.980 $\pm$ 0.001 & 0.983 $\pm$ 0.001 & 0.987 $\pm$ 0.001 & 0.987 $\pm$ 0.001 & 0.988 $\pm$ 0.001 \\
        & 0.90 & 0.908 $\pm$ 0.002 & 0.981 $\pm$ 0.002 & 0.985 $\pm$ 0.001 & 0.986 $\pm$ 0.001 & 0.986 $\pm$ 0.001 \\
        & 0.98 & \textbf{0.982 $\pm$ 0.001} & 0.982 $\pm$ 0.001 & 0.981 $\pm$ 0.001 & 0.981 $\pm$ 0.001 & 0.981 $\pm$ 0.001 \\
        \bottomrule
    \end{tabular}
    }
       \label{table:masking_color_std}
\end{table*}

Additionally, we demonstrate that the context mask is essential for removing shortcuts in the invariance setting. The performance comparison between SimCLR$^+$ and the model without the context mask is presented in \Cref{table:ablation_inv} for 3DIEBench and \Cref{table:ablation_inv_cifar} for CIFAR10. Without the context mask, classification accuracy drops significantly, highlighting its crucial role.

\begin{table*}[!htb]
    \centering
    \caption{Performance of the invariant SimCLR$^+$ on equivariant (rotation prediction, color prediction) and invariant (classification) tasks in 3DIEBench. Both relative value predictions and individual latent value predictions are reported for prediction tasks. Context masking is essential for invariant model performance.}
    \resizebox{.8\textwidth}{!}{%
    \begin{tabular}{lccccc}
        \toprule
        Method & \multicolumn{2}{c}{Rotation prediction ($R^2$)} & \multicolumn{2}{c}{Color prediction ($R^2$)}  & \multicolumn{1}{c}{Classification (top-1)}  \\
        \cmidrule(lr){2-3} \cmidrule(lr){4-5} \cmidrule(lr){6-6}
        & Relative & Individual & Relative & Individual & Representation \\
        \midrule
        SimCLR$^+$ & 0.489 & 0.773 & 0.130 & 0.116 & 81.0 \\
        SimCLR$^+$ (w/o mask) & 0.247 & 0.544 & 0.464 & 0.498 & 42.3 \\
        \bottomrule
    \end{tabular}
    }
    \label{table:ablation_inv}
\end{table*}

\begin{table*}[!htb]
    \centering
    \caption{Performance of the invariant SimCLR$^+$ on equivariant (crop prediction, blur prediction, color prediction) and invariant (classification) tasks in CIFAR-10. Both relative value predictions and individual latent value predictions are reported for prediction tasks. Context masking is essential for invariant model performance.}
    \resizebox{1.\textwidth}{!}{%
    \begin{tabular}{lccccccc}
        \toprule
        Method & \multicolumn{2}{c}{Crop prediction ($R^2$)} & \multicolumn{2}{c}{Blur prediction ($R^2$)} & \multicolumn{2}{c}{Color prediction ($R^2$)}  & \multicolumn{1}{c}{Classification (top-1)}  \\
        \cmidrule(lr){2-3} \cmidrule(lr){4-5} \cmidrule(lr){6-7} \cmidrule(lr){8-8}
        & Relative & Individual & Relative & Individual & Relative & Individual & Representation \\
        \midrule
        SimCLR$^+$ & 0.505 & 0.453 & 0.381 & 0.170 & 0.121 & 0.103 & 89.7 \\
        SimCLR$^+$ (w/o mask) & 0.362 & 0.202 & 0.444 & 0.322 & 0.318 & 0.242 & 59.9 \\
        \bottomrule
    \end{tabular}
    }
    \label{table:ablation_inv_cifar}
\end{table*}

\subsubsection{Role of Auxiliary Predictor}

We provide the complete results corresponding to \Cref{fig:ablation_aug_pred} in \Cref{table:auxiliary} to demonstrate that the auxiliary predictor is crucial for the model to achieve equivariance. For better comparison, the rotation and color prediction performance for the invariance baselines (SimCLR, SimCLR$^+$(c=0), SimCLR$^+$) is based on the predictor output instead of the features.
In its absence, while the model retains its performance on the invariant classification task, it fails to learn equivariance, performs similarly to the invariant models, and cannot effectively adapt to different contexts.

\begin{table*}[!htb]
    \centering
    \caption{Performance of \algo on classification, rotation and color prediction tasks in 3DIEBench with and without the auxiliary predictor}
    \resizebox{1.\textwidth}{!}{%
    \begin{tabular}{lccccccccccc}
        \toprule
        Method & \multicolumn{5}{c}{Rotation prediction ($R^2$)} & \multicolumn{5}{c}{Color prediction ($R^2$)} & \multicolumn{1}{c}{Classification (top-1)}  \\
        \cmidrule(lr){2-6} \cmidrule(lr){7-11}\cmidrule(lr){12-12}
        & 0 & 2 & 14 & 30 & 126 & 0 & 2 & 14 & 30 & 126 & Representation \\
        \midrule
        SimCLR  & \multicolumn{5}{c}{0.227} & \multicolumn{5}{c}{-0.004} & 85.3  \\
        SimCLR$^+$ (c=0)& \multicolumn{5}{c}{0.230} & \multicolumn{5}{c}{-0.004} & 83.4 \\
        SimCLR$^+$ & \multicolumn{5}{c}{0.228} & \multicolumn{5}{c}{-0.004} & 81.0 \\
        \algo (w/o) (rotation) & 0.227 & 0.227 & 0.226 & 0.226 & 0.227 & -0.003 & -0.003 & -0.003 & -0.004 & -0.004 & 80.8 \\
        \algo (w/o) (color) & 0.227 & 0.227 & 0.226 & 0.226 & 0.227 & -0.003 & -0.003 & -0.003 & -0.004 & -0.004 & 80.8 \\
        \algo (rotation) & 0.734 & 0.740 & 0.743 & 0.743 & 0.744 & 0.908 & 0.664 & 0.037 & 0.023 & 0.046 & 80.4 \\
        \algo (color) & 0.735 & 0.614 & 0.389 & 0.345 & 0.344 & 0.908 & 0.981 & 0.985 & 0.986 & 0.986 & 80.4 \\
        \bottomrule
    \end{tabular}
    }
       \label{table:auxiliary}
\end{table*}


\subsection{Qualitative Assessment of Adaptation to Task-Specific Symmetries}\label{sec:knn_supplement}

\subsubsection{Comparison with Baseline Approaches}

\begin{figure}[!htb]
    \centering
    \includegraphics[width=0.8\textwidth]{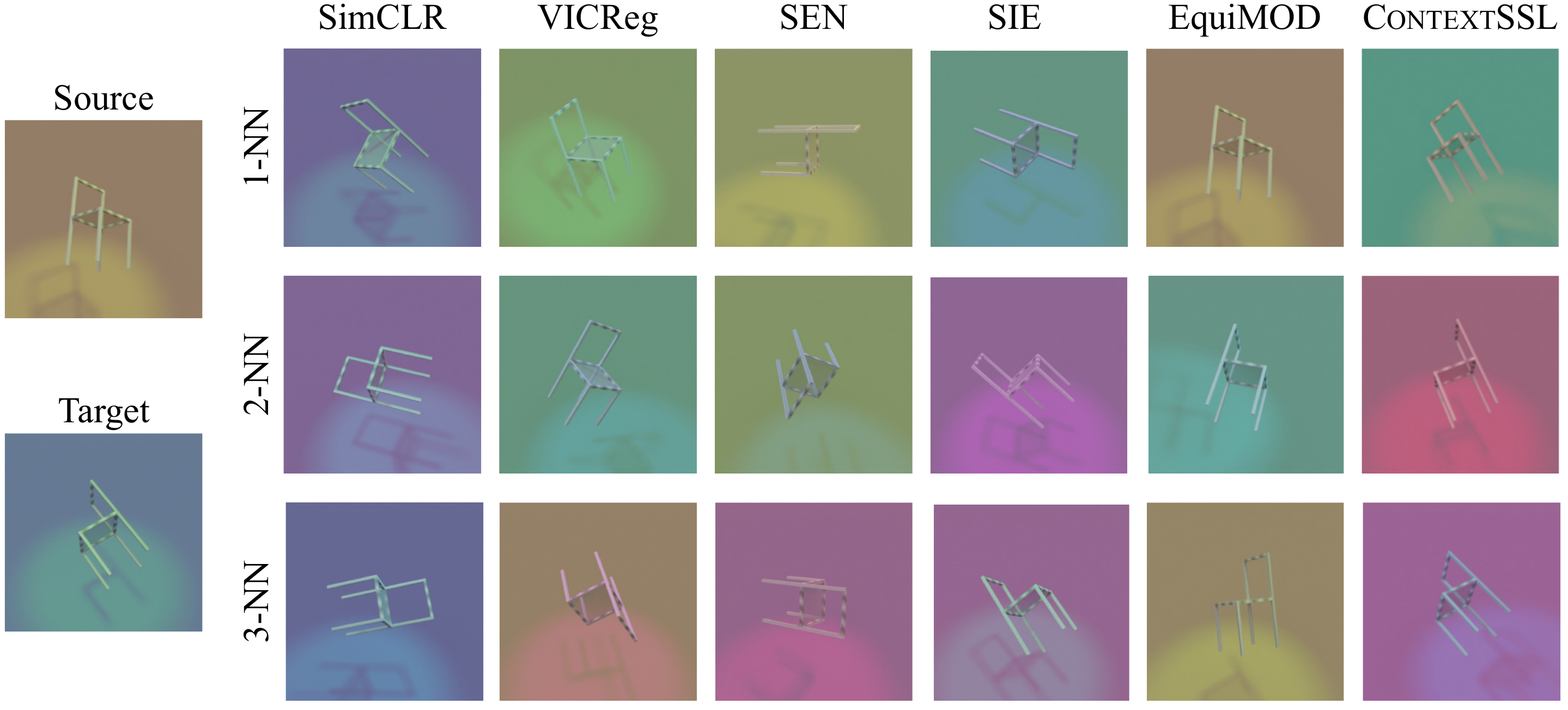}
    \caption{Nearest neighbors of different methods taking as input the source image and rotation angle. \algo aligns best with the rotation angle of the target image.}
    \label{fig:knn_baselines_supplement}
\end{figure}

We provide additional results to the qualitative assessment comparing with different models in \Cref{fig:knn_baselines_supplement}.
The nearest neighbors of invariance models (SimCLR and VICReg) have random rotation angles. Equivariance baselines (SEN, SIE, EquiMOD) correctly generate the target rotation angle for some of the 3-nearest neighbors but fail in others. \algo outperforms by successfully identifying the correct angle in all 3-nearest neighbors while remaining invariant to color variations.

\subsubsection{Nearest Neighbour Retrieval with Varying Context}

\begin{figure}[!htb]
    \centering
    \includegraphics[width=1.\textwidth]{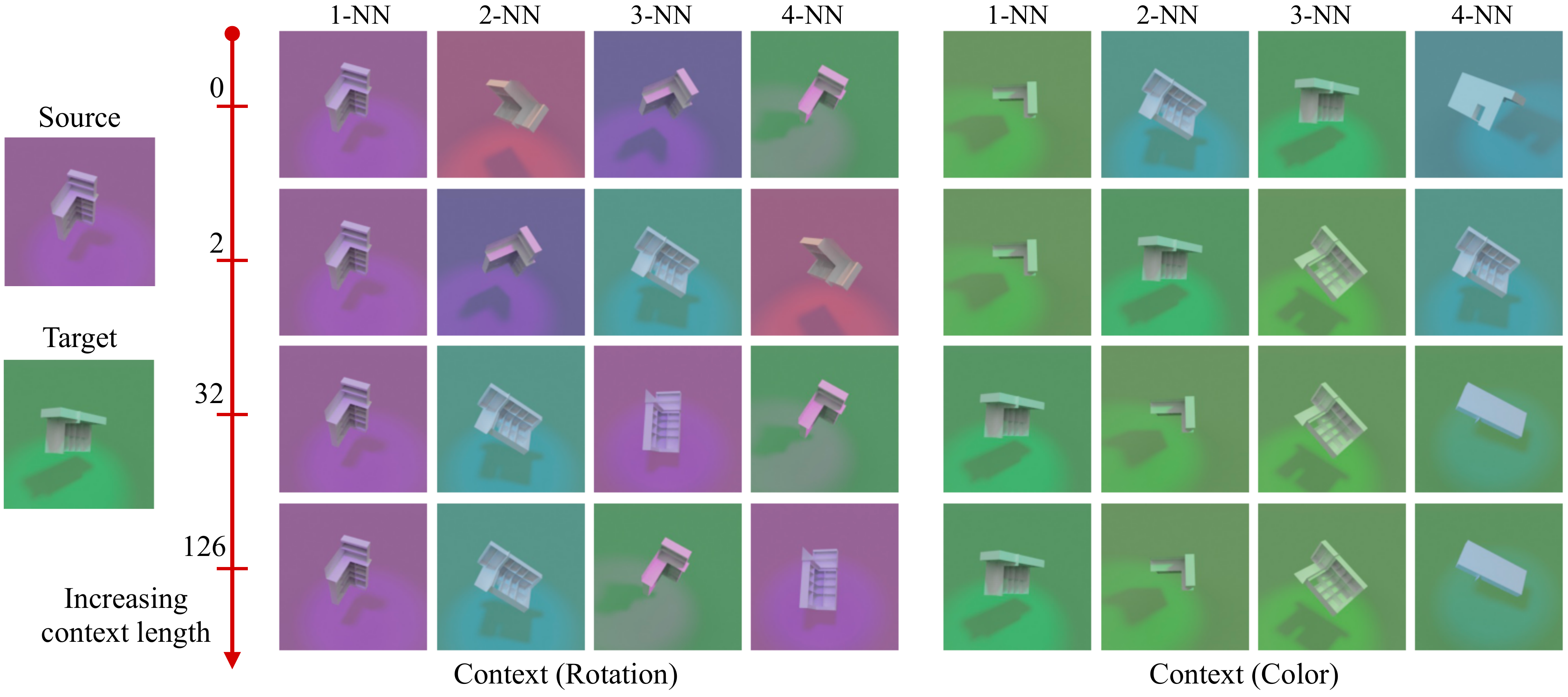}
    \caption{Nearest neighbors of \algo taking as input the source image and rotation angle at different context lengths. As context increases, \algo aligns better with the rotation angle (color) of the target image when the context is based on rotation (color).}
    \label{fig:knn_with_context_v1}
\end{figure}

\begin{figure}[!htb]
    \centering
    \includegraphics[width=1.\textwidth]{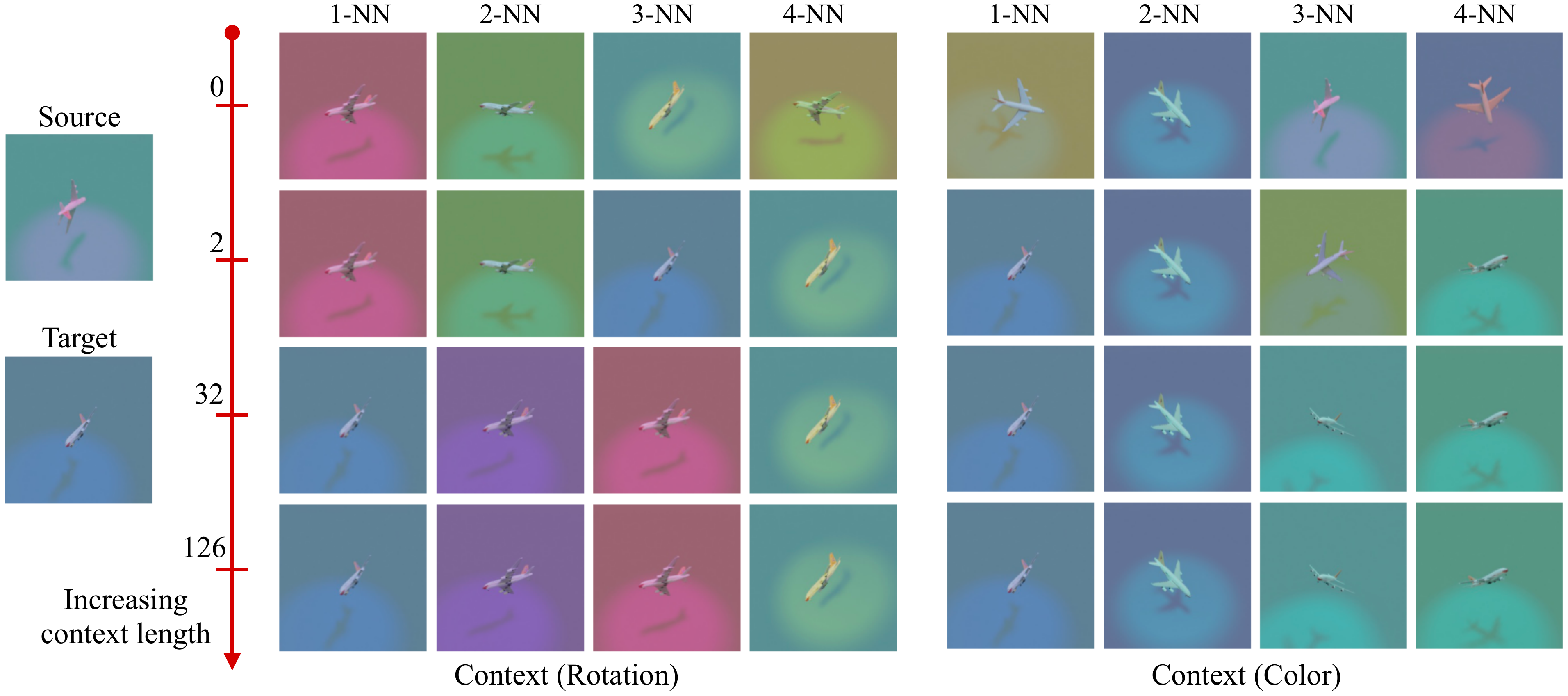}
    \caption{Nearest neighbors of \algo taking as input the source image and rotation angle at different context lengths. As context increases, \algo aligns better with the rotation angle (color) of the target image when the context is based on rotation (color).}
    \label{fig:knn_with_context_v2}
\end{figure}

\begin{figure}[!htb]
    \centering
    \includegraphics[width=1.\textwidth]{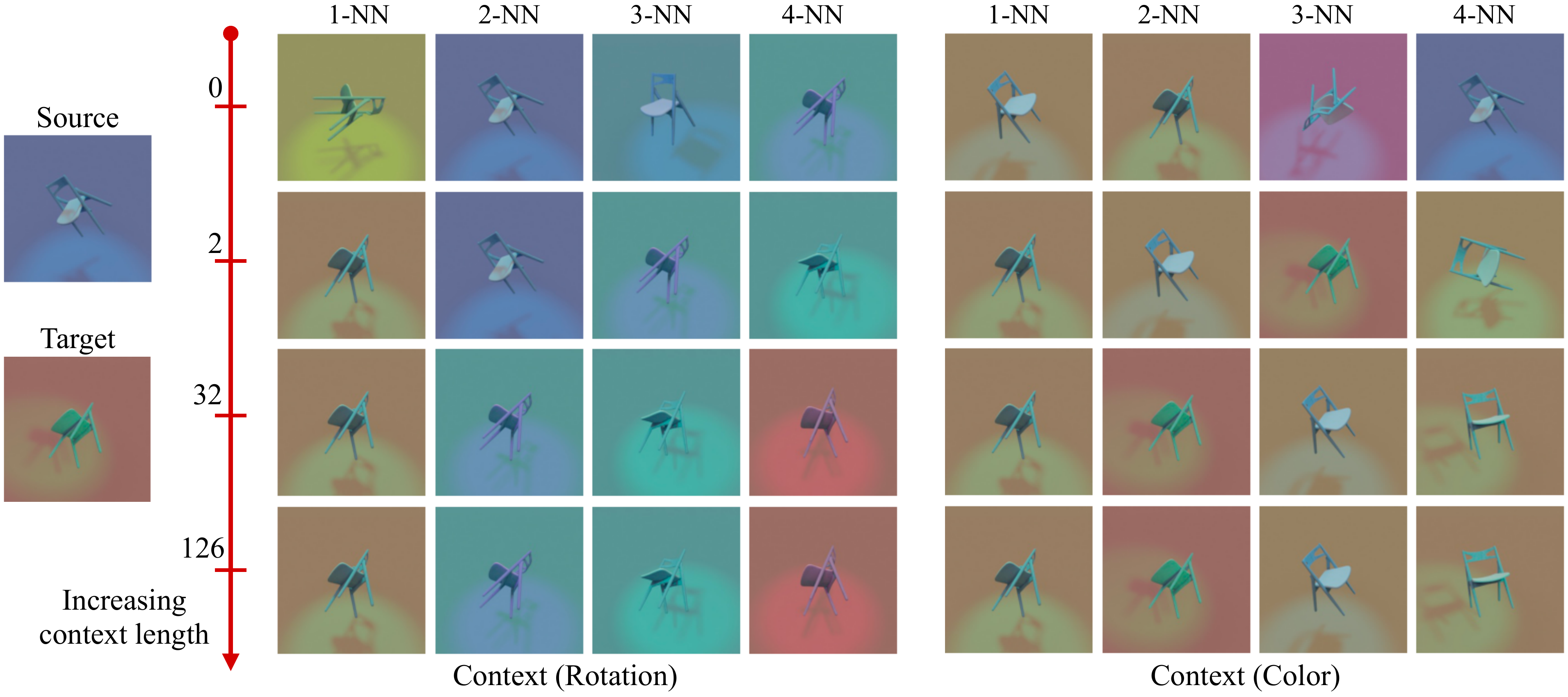}
    \caption{Nearest neighbors of \algo taking as input the source image and rotation angle at different context lengths. As context increases, \algo aligns better with the rotation angle (color) of the target image when the context is based on rotation (color).}
    \label{fig:knn_with_context_v3}
\end{figure}

In this section, we conduct a qualitative assessment of model performance by taking the nearest neighbors of the predictor output when inputting a source image and a transformation variable, and show the change in retrieving quality in \Cref{fig:knn_with_context_v1}, \Cref{fig:knn_with_context_v2}, and \Cref{fig:knn_with_context_v3}. We observe that the nearest neighbors have a closer rotation angle (color) to the target image under rotation (color) context as context length increases, indicating \algo's ability to adapt to the given context as context length increases.

\subsection{Expanding to Diverse Data Transformations}\label{sec:supplement_cifar10_all}

Unlike 3DIEBench where meta-latents for each data are available, we manually construct positives by applying augmentations like crop and blur on CIFAR10. The results for the combinations of crop and blur are reported in~\Cref{table:cifar10_results_cropblur}.
We additionally provide the results for the combinations of crop and color in \Cref{table:cifar10_results_cropcolor} and crop and blur in \Cref{table:cifar10_results_cropblur}.
Consistent with our previous results, while almost retaining the classification performance as SimCLR, \algo learns to adaptively enforce equivariance and invariance to different environments depending upon the context.

\begin{table*}[!htb]
    \centering
    \caption{\textbf{CIFAR-10 Color-Blur.} Performance of \algo on invariant (classification) and equivariant (color prediction, blur prediction) tasks in CIFAR-10 under the environment of color, i.e. \algo (color), and blur, i.e. \algo (blur).}
    \resizebox{1.\textwidth}{!}{%
    \begin{tabular}{lccccccccccc}
        \toprule
        Method & \multicolumn{5}{c}{Color prediction ($R^2$)} & \multicolumn{5}{c}{Blur prediction ($R^2$)} & \multicolumn{1}{c}{Classification (top-1)}  \\
        \cmidrule(lr){2-6} \cmidrule(lr){7-11}\cmidrule(lr){12-12}
        & 0 & 2 & 14 & 30 & 126 & 0 & 2 & 14 & 30 & 126 & Representation \\
        \midrule
        SimCLR  & \multicolumn{5}{c}{0.154} & \multicolumn{5}{c}{0.371} & 89.1  \\
        SimCLR$^+$ (c=0)&  \multicolumn{5}{c}{0.054} & \multicolumn{5}{c}{0.361} & 88.9  \\
        SimCLR$^+$ & \multicolumn{5}{c}{0.121} & \multicolumn{5}{c}{0.381} & 89.7 \\
        \algo (color) & 0.518 & 0.519 & 0.519 & 0.519 & 0.519 & 0.916 & 0.793 & 0.699 & 0.735 & 0.823 & 88.9 \\
        \algo (blur) & 0.518 & 0.353 & 0.241 & 0.259 & 0.333 & 0.916 & 0.916 & 0.916 & 0.916 & 0.917 & 88.8 \\
        \bottomrule
    \end{tabular}
    }
       \label{table:cifar10_results_colorblur}
\end{table*}

\begin{table*}[!htb]
    \centering
    \caption{\textbf{CIFAR-10 Crop-Color.} Performance of \algo on invariant (classification) and equivariant (crop prediction, color prediction) tasks in CIFAR-10 under the environment of crop, i.e. \algo (crop), and color, i.e. \algo (color).}
    \resizebox{1.\textwidth}{!}{%
    \begin{tabular}{lccccccccccc}
        \toprule
        Method & \multicolumn{5}{c}{Crop prediction ($R^2$)} & \multicolumn{5}{c}{Color prediction ($R^2$)} & \multicolumn{1}{c}{Classification (top-1)}  \\
        \cmidrule(lr){2-6} \cmidrule(lr){7-11}\cmidrule(lr){12-12}
        & 0 & 2 & 14 & 30 & 126 & 0 & 2 & 14 & 30 & 126 & Representation \\
        \midrule
        SimCLR  & \multicolumn{5}{c}{0.459} & \multicolumn{5}{c}{0.154} & 89.1  \\
        SimCLR$^+$ (c=0)& \multicolumn{5}{c}{0.448} & \multicolumn{5}{c}{0.054} & 88.9  \\
        SimCLR$^+$ & \multicolumn{5}{c}{0.505} & \multicolumn{5}{c}{0.121} & 89.7 \\
        \algo (crop) & 0.606 & 0.606 & 0.607 & 0.607 & 0.607 & 0.522 & 0.378 & 0.253 & 0.264 & 0.301 & 87.5 \\
        \algo (color) & 0.605 & 0.467 & 0.387 & 0.466 & 0.511 & 0.523 & 0.525 & 0.527 & 0.527 & 0.527 & 87.5\\
        \bottomrule
    \end{tabular}
    }
       \label{table:cifar10_results_cropcolor}
\end{table*}

In addition to the results for predicting relative latent values between pairs of image embeddings as input in \Cref{table:cifar10_results_cropblur}, \Cref{table:cifar10_results_cropcolor}, and \Cref{table:cifar10_results_colorblur}, we provide the evaluation of learned representations on equivariant tasks (rotation and color prediction) to predict individual latent values, as shown in \Cref{table:cifar10_results_cropblur_indi}, \Cref{table:cifar10_results_cropcolor_indi}, and \Cref{table:cifar10_results_colorblur_indi} respectively. Both results lead to the same conclusion, that \algo is able to adaptively enforce equivariance and invariance to different environments depending upon the context.

\begin{table*}[!htb]
    \centering
    \caption{\textbf{CIFAR-10 Crop-Blur.} Performance of \algo on equivariant (crop prediction, blur prediction) tasks in CIFAR-10 under the environment of crop, i.e. \algo (crop), and blur, i.e. \algo (blur), to predict individual latent values.}
    \resizebox{1.\textwidth}{!}{%
    \begin{tabular}{lcccccccccc}
        \toprule
        Method & \multicolumn{5}{c}{Crop prediction ($R^2$)} & \multicolumn{5}{c}{Blur prediction ($R^2$)} \\
        \cmidrule(lr){2-6} \cmidrule(lr){7-11}
        & 0 & 2 & 14 & 30 & 126 & 0 & 2 & 14 & 30 & 126 \\
        \midrule
        SimCLR  & \multicolumn{5}{c}{0.382} & \multicolumn{5}{c}{0.122} \\
        SimCLR$^+$ (c=0)& \multicolumn{5}{c}{0.375} & \multicolumn{5}{c}{0.111} \\
        SimCLR$^+$ & \multicolumn{5}{c}{0.453} & \multicolumn{5}{c}{0.170} \\
        \algo (crop) & 0.576 & 0.575 & 0.576 & 0.576 & 0.576 & 0.835 & 0.795 & 0.630 & 0.644 & 0.663  \\
        \algo (blur) & 0.575 & 0.504 & 0.463 & 0.443 & 0.474 & 0.835 & 0.835 & 0.836 & 0.837 & 0.837  \\
        \bottomrule
    \end{tabular}
    }
       \label{table:cifar10_results_cropblur_indi}
\end{table*}

\begin{table*}[!htb]
    \centering
    \caption{\textbf{CIFAR-10 Color-Blur.} Performance of \algo on equivariant (color prediction, blur prediction) tasks in CIFAR-10 under the environment of color, i.e. \algo (color), and blur, i.e. \algo (blur), to predict individual latent values.}
    \resizebox{1.\textwidth}{!}{%
    \begin{tabular}{lccccccccccc}
        \toprule
        Method & \multicolumn{5}{c}{Color prediction ($R^2$)} & \multicolumn{5}{c}{Blur prediction ($R^2$)} \\
        \cmidrule(lr){2-6} \cmidrule(lr){7-11}\cmidrule(lr){12-12}
        & 0 & 2 & 14 & 30 & 126 & 0 & 2 & 14 & 30 & 126 \\
        \midrule
        SimCLR  & \multicolumn{5}{c}{0.121} & \multicolumn{5}{c}{0.122}  \\
        SimCLR$^+$ (c=0)&  \multicolumn{5}{c}{0.039} & \multicolumn{5}{c}{0.111}   \\
        SimCLR$^+$ & \multicolumn{5}{c}{0.103} & \multicolumn{5}{c}{0.170} \\
        \algo (color) & 0.488 & 0.488 & 0.488 & 0.488 & 0.488 & 0.837 & 0.711 & 0.628 & 0.672 & 0.730 \\
        \algo (blur) & 0.488 & 0.376 & 0.286 & 0.309 & 0.362 & 0.837 & 0.838 & 0.838 & 0.838 & 0.837 \\
        \bottomrule
    \end{tabular}
    }
       \label{table:cifar10_results_colorblur_indi}
\end{table*}

\begin{table*}[!htb]
    \centering
    \caption{\textbf{CIFAR-10 Crop-Blur.} Performance of \algo on equivariant (crop prediction, color prediction) tasks in CIFAR-10 under the environment of crop, i.e. \algo (crop), and color, i.e. \algo (color), to predict individual latent values.}
    \resizebox{1.\textwidth}{!}{%
    \begin{tabular}{lcccccccccc}
        \toprule
        Method & \multicolumn{5}{c}{Crop prediction ($R^2$)} & \multicolumn{5}{c}{Color prediction ($R^2$)} \\
        \cmidrule(lr){2-6} \cmidrule(lr){7-11}
        & 0 & 2 & 14 & 30 & 126 & 0 & 2 & 14 & 30 & 126 \\
        \midrule
        SimCLR  & \multicolumn{5}{c}{0.382} & \multicolumn{5}{c}{0.121} \\
        SimCLR$^+$ (c=0)& \multicolumn{5}{c}{0.375} & \multicolumn{5}{c}{0.039} \\
        SimCLR$^+$ & \multicolumn{5}{c}{0.453} & \multicolumn{5}{c}{0.103} \\
        \algo (crop) & 0.570 & 0.572 & 0.572 & 0.572 & 0.572 & 0.495 & 0.417 & 0.342 & 0.356 & 0.373 \\
        \algo (color) & 0.570 & 0.490 & 0.447 & 0.492 & 0.515 & 0.495 & 0.496 & 0.497 & 0.497 & 0.497 \\
        \bottomrule
    \end{tabular}
    }
       \label{table:cifar10_results_cropcolor_indi}
\end{table*}

\subsection{Context World Models Beyond Self-Supervised Learning}\label{sec:supervised_supplement}

We report classification performance along with rotation and color prediction equivariant measures. The results for predicting relative values are shown in \Cref{table:main_supervised_table} and the results for predicting individual latent values are shown in \Cref{table:rotation_supervised_table}. The equivariance (invariance) performance of \algo improves with increased context.

\begin{table*}[!htb]
    \centering
    \caption{\textbf{Context-Dependent Labels Classification Task.} Performance of \algo on equivariant (rotation prediction, color prediction) tasks for context-dependent labels to predict individual latent values. As context length increases, \algo becomes more equivariant to color (or rotation) and more invariant to rotation (or color) within the respective environment.}
    \resizebox{1.\textwidth}{!}{%
    \begin{tabular}{lcccccccccc}
        \toprule
        Method & \multicolumn{5}{c}{Rotation prediction ($R^2$)} & \multicolumn{5}{c}{Color prediction ($R^2$)} \\
        \cmidrule(lr){2-6} \cmidrule(lr){7-11}
        & 0 & 2 & 14 & 30 & 126 & 0 & 2 & 14 & 30 & 126 \\
        \midrule
        SimCLR & \multicolumn{5}{c}{0.781} & \multicolumn{5}{c}{0.058}  \\
        SimCLR$^+$ (c=0) & \multicolumn{5}{c}{0.478} & \multicolumn{5}{c}{-0.003} \\
        SimCLR$^+$ & \multicolumn{5}{c}{0.695} & \multicolumn{5}{c}{0.267}  \\
        \algo (color) & 0.751 & 0.751 & 0.750 & 0.750 & 0.749 & 0.915 & 0.973 & 0.980 & 0.981 & 0.981 \\
        \algo (rotation) & 0.750 &  0.778 & 0.797 & 0.795 & 0.795 & 0.915 & 0.375 & 0.104 & 0.091 & 0.090  \\
        \bottomrule
    \end{tabular}
    }
       \label{table:rotation_supervised_table}
\end{table*}

\subsection{Performance on Encoder Representations and Predictor Embedding}\label{sec: predictor_vs_rep_supplement}

We analyze the difference between the performance on representation and the performance on predictor embedding for both the invariance (classification) task and equivariance (rotation prediction) task in \Cref{table:repr_emb_compare} and \Cref{table:predictor_acc}. \algo maintains almost the same performance for rotation prediction using either representations or embeddings, while the performance of all other baselines drops significantly when using the embeddings. Similar conclusions apply to the classification case.

\begin{table*}[!htb]
    \centering
    \caption{Model performance in rotation prediction task, within the rotation-equivariant environment. The $R^2$ values are calculated for both the representations and the embeddings (output of projection head for invariant models (VICReg, SimCLR) or predictor for equivariant models (SEN, EquiMod, SIE, \algo). Unlike other models, which experience a significant performance drop between representations and embeddings, \algo maintains consistent performance.}
    \resizebox{.7\textwidth}{!}{%
    \begin{tabular}{lccccc}
        \toprule
        Method & \multicolumn{3}{c}{Rotation prediction ($R^2$)} \\
        \cmidrule(lr){2-4} 
        & Representations & Embeddings & Change \\
        \midrule
        VICReg  & 0.37 & 0.23 & -0.14 \\
        SimCLR& 0.51 &0.23 & -0.28\\
        SEN & 0.63 &0.39 & -0.24 \\
        EquiMod & 0.51 & 0.39 & -0.12 \\
        SIE & 0.67 & 0.60 & -0.07 \\
        \algo (rotation) & \textbf{0.74} & \textbf{0.74} & \textbf{-0.00}  \\
        \bottomrule
    \end{tabular}
    }
       \label{table:repr_emb_compare}
\end{table*}

\begin{table*}[!htb]
    \centering
    \caption{Performance of \algo on the accuracy of predictor embeddings for context-dependent labels.}
    \resizebox{.85\textwidth}{!}{%
    \begin{tabular}{lcccccccc}
        \toprule
        Method & \multicolumn{6}{c}{Classification (top-1)} &  \\
        \cmidrule(lr){2-6} \cmidrule(lr){6-7} \cmidrule(lr){8-8} 
        & 0 & 2 & 14 & 30 & 126 & Representation & Change \\
        \midrule
        SimCLR  & \multicolumn{5}{c}{52.7} & 85.3 & -32.6 \\
        SimCLR$^+$ (c=0)& \multicolumn{5}{c}{72.4} & 83.4 & -11.0 \\
        SimCLR$^+$ & \multicolumn{5}{c}{77.8} & 81.0 & -3.2 \\
        \algo (rotation) & 76.6 & 76.9 & 75.6 & 76.9 & 77.5  & 80.4 & \textbf{-2.9} \\
        \algo (color) & 76.6 & 75.3 & 71.7 & 72.6 & 76.5 & 80.4 & -3.9\\
        \bottomrule
    \end{tabular}
    }
       \label{table:predictor_acc}
\end{table*}


\subsection{Enforcing Invariance or Equivariance to the Same Transformation Using Context}\label{sec:single env}
Apart from adaptively learning equivariance to a subset of transformation groups and invariance to the rest as shown in~\Cref{table:main_table}, we extend \algo to operate within environments characterized by a single transformation. Motivated by this, we ask the question: \emph{Can \algo adapt to learn equivariance or invariance to the same transformation depending on the context?}. At training, we randomly sample one of these environments. If the environment corresponds to enforcing equivariance, we construct our context in the same way as before i.e. pairs of positives transformed using augmentations sampled from the transformation group. However, if the environment corresponds to enforcing invariance, we maximize alignment between positives transformed by augmentation sampled from the transformation group without conditioning on that augmentation. Take rotation in 3DIEBench as an example. As shown in~\Cref{table:single_environment}, similar to our results in two transformation setting (rotation and color) in \Cref{table:main_table}, \algoname effectively adapts to enforce invariance and equivarance to rotation depending on the context. Results for predicting individual latents are provided in \Cref{table:single_environment_indi}.

\begin{table*}[!htb]
    \centering
    \caption{\textbf{Single Transformation Setting.} Performance of \algo in 3DIEBench under the equivariant environment, i.e. \algo (rotation), and the invariant environment, i.e. \algo (none), with respect to rotation.}
    \resizebox{.7\textwidth}{!}{%
    \begin{tabular}{lccccccc}
        \toprule
        Method & \multicolumn{5}{c}{Rotation prediction ($R^2$)}  & \multicolumn{1}{c}{Classification (top-1)}  \\
        \cmidrule(lr){2-6} \cmidrule(lr){7-7}
        & 0 & 2 & 14 & 30 & 126 & Representation \\
        \midrule
        SimCLR  & \multicolumn{5}{c}{0.506} & 85.3 \\
        SimCLR$^+$ (c=0)& \multicolumn{5}{c}{0.478} & 83.4 \\
        SimCLR$^+$ & \multicolumn{5}{c}{0.489} & 81.0 \\
        \algo (rotation) & 0.737 & 0.737 & 0.736 & 0.737 & 0.738 & 80.6 \\
        \algo (none) & 0.737 & 0.717 & 0.477 & 0.377 & 0.473 & 80.6 \\
        \bottomrule
    \end{tabular}
    }
       \label{table:single_environment}
\end{table*}

\begin{table*}[!htb]
    \centering
    \caption{\textbf{Single Transformation Setting.} Performance of \algo in 3DIEBench under the equivariant environment, i.e. \algo (rotation), and the invariant environment, i.e. \algo (none), with respect to rotation, to predict the individual latent values.}
    \resizebox{.6\textwidth}{!}{%
    \begin{tabular}{lccccccc}
        \toprule
        Method & \multicolumn{5}{c}{Rotation prediction ($R^2$)} \\
        \cmidrule(lr){2-6} 
        & 0 & 2 & 14 & 30 & 126 \\
        \midrule
        SimCLR  & \multicolumn{5}{c}{0.791} \\
        SimCLR$^+$ (c=0)& \multicolumn{5}{c}{0.773}  \\
        SimCLR$^+$ & \multicolumn{5}{c}{0.773} \\
        \algo (rotation) & 0.778  & 0.777 & 0.767  & 0.768 & 0.777   \\
        \algo (none) & 0.839 & 0.829 & 0.721 & 0.667 & 0.698  \\
        \bottomrule
    \end{tabular}
    }
       \label{table:single_environment_indi}
\end{table*}